\documentclass{article}
\usepackage[final]{colm2026_conference}

\usepackage{microtype}
\usepackage{graphicx}
\usepackage{booktabs} 
\usepackage[T1]{fontenc}
\usepackage{listings}
\usepackage[dvipsnames]{xcolor}

\lstdefinestyle{pythonstyle}{
  language=Python,
  basicstyle=\ttfamily\scriptsize,
  keywordstyle=\color[rgb]{0.13,0.29,0.53}\bfseries,
  emphstyle=\color[rgb]{0.13,0.29,0.53},
  stringstyle=\color[rgb]{0.31,0.54,0.16},
  commentstyle=\color[rgb]{0.56,0.56,0.56},
  numberstyle=\tiny\color{gray},
  numbers=left, numbersep=5pt,
  frame=single, framesep=3pt,
  rulecolor=\color{gray!40},
  backgroundcolor=\color{gray!4},
  breaklines=true, showstringspaces=false, tabsize=4,
  xleftmargin=1.5em, framexleftmargin=1em,
  columns=flexible,
}

\lstdefinestyle{wrappedverbatim}{
  basicstyle=\ttfamily\footnotesize,
  breaklines=true,
  breakatwhitespace=true,
  columns=fullflexible,
  keepspaces=true,
  showstringspaces=false,
}

\usepackage{lineno}

\usepackage{hyperref}
\usepackage{xurl}

\definecolor{darkblue}{rgb}{0, 0, 0.5}
\hypersetup{colorlinks=true, citecolor=darkblue, linkcolor=darkblue, urlcolor=darkblue}
\usepackage{float}

\usepackage{amsmath}

\usepackage[capitalize,noabbrev]{cleveref}

\title{Recovering Wasted Compute in Autoresearch Agents}

\author{Au Kwok Chun\textsuperscript{\rm 1$\ast$},
  Abhigyan Acherjee\textsuperscript{\rm 2$\ast$},
  Amrutha Rao\textsuperscript{\rm 3$\ast$}, \\
  \textbf{Zaiqian Chen\textsuperscript{\rm 4},
    Kazem Meidani\textsuperscript{\rm 5},
    C. Bayan Bruss\textsuperscript{\rm 5},
  Micah Goldblum\textsuperscript{\rm 1, 6}}
}

\begin{document}

\ifcolmsubmission
\linenumbers
\fi

\maketitle
\let\thefootnote\relax\footnotetext{
  \begin{minipage}{\dimexpr\textwidth-1.8em\relax}
    \textsuperscript{$\ast$}Equal contribution. \\
    \textsuperscript{\rm 1}Department of Computer Science, Columbia University \quad
    \textsuperscript{\rm 2}AI, Analytics and Future of Work Initiative, Georgetown University \quad
    \textsuperscript{\rm 3}Department of Applied Mathematics, Columbia University \quad
    \textsuperscript{\rm 4}Department of Statistics, Columbia University \quad
    \textsuperscript{\rm 5}Capital One \quad
    \textsuperscript{\rm 6}Department of Electrical Engineering, Columbia University \\
    \vspace{1ex} \\
    Correspondence to: \texttt{ka3094@columbia.edu, arr2249@columbia.edu, aa3320@georgetown.edu}
  \end{minipage}
}

\begin{abstract}
  A slew of recent works develop agents for solving research
  problems end-to-end, a paradigm increasingly referred to as autoresearch. Such
  agents have inspired large industry investment, motivated by their potential to
  automate time-consuming human labor and customize machine learning solutions for
  specialized applications. In this paper, we study the modeling pipeline at the core of these autoresearch systems and identify common failure modes when they are
  applied to tabular datasets: (1) they waste compute resolving the same bugs over
  and over again; (2) they often fail to tune hyperparameters even when they have a
  large remaining compute budget; (3) the tree-search algorithms that power them do
  not explore; and (4) they perform data analysis, mimicking the humans whose data
  they are trained on, but do not use that analysis to make downstream decisions. We
  explore targeted interventions and find that a global debug consultant that shares
  discovered runtime constraints across all branches of the search tree, prompt- and
  control-level enhancements, and refined tree-search algorithms successfully recover
  wasted compute. Our results show that large gains in autoresearch agent performance
  are achievable through agentic design alone, holding the underlying language model
  fixed.
\end{abstract}

\section{Introduction}
\label{intro}

Fueled by large-scale industry investment, agentic systems powered by Large Language Models (LLMs) are rapidly replacing traditional AutoML pipelines for automated data science \citep{jing2025dsbenchfardatascience, chan2024mlebench}. Increasingly, these systems are aimed at broader research loops, from generating and verifying hypotheses purely from data \citep{pmlr-v235-majumder24a, majumder2025discoverybench} to running experiments and writing up findings \citep{lu2024aiscientistfullyautomated}, a direction known as \emph{autoresearch}. No matter the goal, these systems all rest on the same core task: writing code to process data, train and evaluate models, and produce a candidate solution. Current systems execute this task unreliably: they frequently time out, waste
compute, and produce suboptimal, generic solutions \citep{toledo2025airesearch,
liu2025mlmaster, yang2025rdagent}.

In this work, we identify several recurring failure modes in leading tree-search based agentic frameworks \citep{jiang2025aide, liu2025mlmaster} when applied to tabular machine learning tasks. Each represents a distinct category of how an agent's compute budget is wasted rather than productively spent. First, agents waste budget rediscovering known bugs. Because branches in a tree search do not share memory of past failures, parallel branches repeatedly resolve identical errors in isolation, preventing meaningful iteration \citep{zhang2025agentic, yin2024thinkrepairselfdirectedautomatedprogram}. Second, agents leave budget on the table by terminating search prematurely. Superficial convergence criteria cause them to stop after only a few valid solutions, skipping the hyperparameter tuning phase that the remaining budget could have supported.
Finally, agents spend budget on unproductive search states, becoming trapped in dead-end solution paths until the budget is exhausted \citep{toledo2025airesearch, pmlr-v235-zhou24r}.

We show that improved agentic design can address these problems without modifying the base LLM. We introduce a suite of structural interventions to resolve them:
\footnote{Code available \href{https://github.com/tingtang2/autoresearch-compute-recovery/tree/main}{here}}

\begin{itemize}
  \item \textbf{Context-aware debugging:} To resolve context isolation, we introduce a debug consultant that enables adaptive learning of the execution environment across the search tree. The consultant accumulates discovered bugs into a shared registry and injects constraints before each generation step, preventing redundant error correction and significantly improving efficiency.
  \item \textbf{Budget-aware hyperparameter tuning enforcement:}
    We identify the absence of systematic hyperparameter optimization as a primary failure mode of autonomous ML agents. To address this, we introduce prompt-level guidance and control-loop-level enforcement mechanisms that compel agents to allocate their compute budgets toward structured hyperparameter tuning. By penalizing local convergence and rewarding validation-driven search, we prevent premature termination and redirect search effort toward fine-grained exploitation.

  \item \textbf{Thompson Sampling-enhanced backtracking:} We replace random backtracking with Monte Carlo Tree Search (MCTS) using Thompson Sampling. This probabilistic approach allows the agent to intelligently navigate the solution space and escape unproductive debugging loops, resulting in a significant improvement in stability of generated solutions.

  \item \textbf{Diagnostic - agents fail to act on injected exploratory data analysis (EDA):} We evaluate whether agents incorporate analytical insights by injecting ``adversarial'' results of a toy exploratory data analysis directly into the context window. Our experiments reveal that current agents tend to ignore these signals, motivating the need for stricter control loops that encourage data-driven planning.
\end{itemize}

\section{Background}
\label{background}

\paragraph{Automated data science.}
The autoresearch systems described above are built on automated data science: the
end-to-end process of solving a data science task without human intervention. The
agents that carry out this process, often called machine learning engineering (MLE)
agents, take a dataset and problem description, generate code to explore and process
the data, train and evaluate candidate models, and return a solution ready for
deployment.

Recent research has shown that LLMs alone cannot solve these open-ended problems effectively and that an agentic scaffold is necessary to guide solution generation \citep{nathani2025mlgymnewframeworkbenchmark}. External tools \citep{toollearning}, execution feedback \citep{gehring2025rlef}, and context management \citep{jiang2025aide, liu2025mlmaster, yang2025rdagent} have all been shown to improve performance. One of the main architectural components in leading MLE agent frameworks \citep{liu2025mlmaster, jiang2025aide} is tree search over the space of candidate programs, where each node represents a solution in the form of a codebase, and edges represent attempted improvements via code edits. At each step, the agent selects a promising node to expand, generates a revised solution, executes it, and uses the resulting validation score to guide further exploration. This iterative execution-in-the-loop design has become the dominant paradigm for agentic data science and forms the basis of the frameworks we study in this work.

\paragraph{Agentic scaffolds.}
The two primary agentic scaffolds we study are AIDE \citep{jiang2025aide} and ML-Master \citep{liu2025mlmaster}. Our choice of these agents is based on their open-source nature and their exceptional performance on MLE-bench~\citep{chan2024mlebench}.
AIDE structures its search around three core components. The first is a deterministic greedy search policy $\pi$ that determines at each step whether to draft a new solution from scratch, debug a buggy node, or improve a valid one. A coding operator $f$ implements these three actions, each with specialized prompts: \emph{drafting} produces an initial implementation, \emph{debugging} repairs execution errors, and \emph{improving} proposes a single atomic change to a working solution. Finally, a summarization operator $\Sigma$ extracts concise summaries of past solutions and their scores, keeping the context manageable as the tree grows.

ML-Master extends this paradigm with an improved search strategy and an explicit reasoning mechanism. Its exploration module is inspired by Monte Carlo Tree Search (MCTS): nodes are selected using the Upper Confidence Bound for Trees (UCT) criterion, balancing each node's accumulated reward against its visit count to prioritize promising but under-explored branches. Nodes are expanded via the same draft, debug, and improve actions as AIDE and assigned a reward based on code execution: a node receives a positive reward if its solution is bug-free and improves upon the best validation metric seen so far, and a negative reward otherwise. Multiple workers explore branches in parallel, with rewards backpropagated through the tree to guide subsequent selection. In place of AIDE's summarization operator, ML-Master employs a reasoning module that embeds a curated memory of past execution results and sibling node insights directly into the reasoning component of the LLM, enabling the agent to learn across parallel exploration paths and leverage the capabilities of reasoning models.

\section{Methodology}
\label{sec:method}

\subsection{Context-aware debug consultant}
\label{sec:context_scaffold}

\begin{figure}[t]
  \centering
  \includegraphics[width=\linewidth]{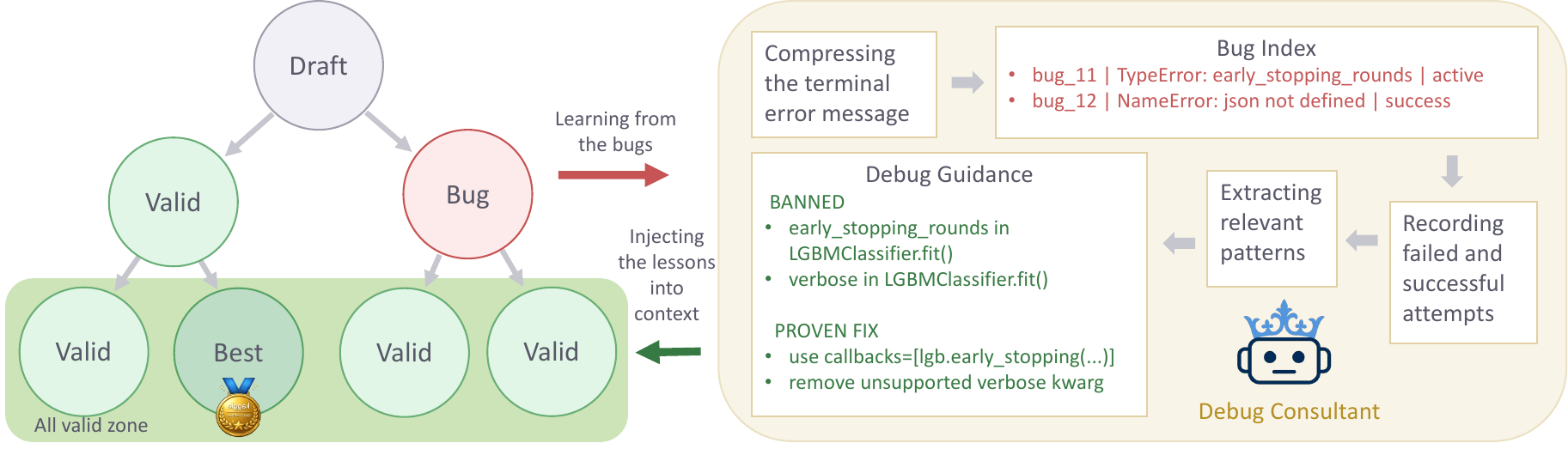}
  \caption{\textbf{The debug consultant shares context across the search tree: one node's failure becomes every node's lesson.} Crashed nodes are compressed into a shared bug index of failed and successful repairs, distilled into banned patterns and proven fixes, and injected into every subsequent generation step, so the agent learns its runtime environment once instead of rediscovering each bug per branch.}
  \label{fig:debug-consultant}
\end{figure}

In the tree-search paradigm described in \Cref{background}, knowledge of failures remains local: in both AIDE and ML-Master, only the child spawned to debug a failed node sees the actual error. The rest of the tree therefore independently rediscovers the same deprecated API call or version mismatch, often dozens of times within a single run.
We call this context isolation.

Our debug consultant (\Cref{fig:debug-consultant}) addresses this by maintaining a shared registry of runtime constraints---which API calls crash, which alternatives work---and propagating this knowledge to all nodes via a three-step control loop in order to allow adaptive learning of the execution environment:

\textbf{Step 1: Error Compression.}
When a node crashes, the raw traceback is compressed into a compact record: the error type, a short signature, and the strategy that caused the failure.
Raw tracebacks are verbose and vary across iterations; a short hint is sufficient for the LLM to avoid the mistake.
This keeps the context window focused on solution search and ensures the registry scales gracefully~\citep{zhang2025agentic}.

\textbf{Step 2: Shared Bug Registry.}
Each compressed record is accumulated into a shared registry that tracks the error type, which strategies have failed, and---when another node succeeds---the strategy that worked. The system distills a concise list of banned patterns and proven fixes:
\begin{quote}
  \small
  \texttt{BANNED: lgb.train(..., verbose\_eval=N)} $\to$ \texttt{TypeError}\\
  \texttt{USE: callbacks=[lgb.log\_evaluation(period=N)]}
\end{quote}
Every new node immediately inherits all entries discovered so far.

\textbf{Step 3: Constraint Injection.}
Constraints are injected at two levels.
During generation (drafting or improving), the distilled banned-pattern list is appended to the prompt, preventing the agent from repeating known-failing API calls.
During debugging, the injection is more targeted: the system retrieves records relevant to the current error and provides specific failed strategies (marked ``never do this'') along with any proven fixes, enabling informed repair rather than blind guessing.

\textbf{Step 4: Deterministic Control Rules.}
In addition to the shared bug registry, the debug consultant adds deterministic rules to the execution process: execution timeouts and empty logs are treated as terminal dead ends that strictly halt the branch, rather than stochastic noise worth retrying.

\subsection{Hyperparameter tuning interventions}

We test three interventions of increasing invasiveness: a prompt-level directive, a control-loop mechanism that shapes the search reward, and their combination.

\begin{figure}[!h]
  \centering
  \includegraphics[width=\linewidth]{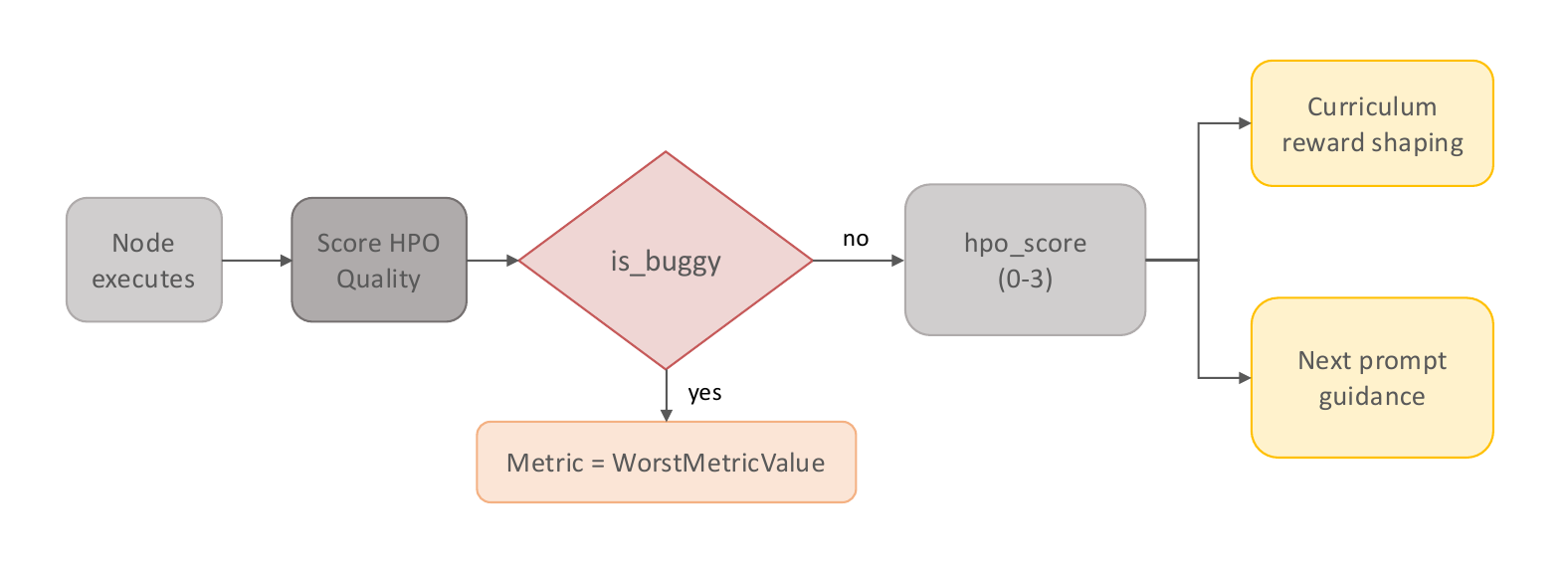}

  \caption{\textbf{Budget dependent control-loop enforcement of hyperparameter tuning in AIDE.} Each node is scored for hyperparameter-tuning quality on a 0--3 scale by an LLM judge. Buggy nodes are assigned the worst possible metric and their score is ignored; for a valid node, the score both conditions the agent's next prompt and adjusts the metric used for node selection. The reward depends on how much budget remains: weak tuning (scores of 0 or 1) is penalized throughout, but strong tuning (2 or 3) is rewarded only in later stages of the search, so the agent explores broadly early and tunes intensively late.}
  \label{fig:hpo-control-loop}
\end{figure}

The \textbf{prompt-level intervention} appends hyperparameter-tuning instructions to the agent context via \texttt{additional\_notes.txt} (for both AIDE and ML-Master), instructing it to establish a validated baseline, run cheap trials to identify the most impactful hyperparameters, then tune those around the best configuration once gains stall.

The \textbf{control-loop intervention} enforces tuning through the search reward rather than the prompt. After a node executes, an execution-time checker (\texttt{\_score\_hyperparameter\_tuning}) grades its tuning quality on a discrete $\{0,1,2,3\}$ scale via an LLM rubric (NONE, MINIMAL, MODERATE, EXTENSIVE). If the node is not buggy, this score biases which nodes the agent expands next.

How the score enters the search differs by agent: in AIDE it adjusts the validation metric and conditions subsequent improvement prompts, while in ML-Master it is folded into the reward used for UCT-based node selection (e.g., $+0.25 \times \texttt{hpo\_score}$).
The AIDE adjustment also accounts for code diversity and parameter reuse: $\text{metric}_{\text{adj}} = \text{metric}_{\text{base}} + 0.1 \times s \times (r_{\text{hpo}} + r_{\text{div}} + r_{\text{corr}})$, where $r_{\text{hpo}}$, $r_{\text{div}}$, and $r_{\text{corr}}$ are the tuning, diversity, and reuse rewards. The scale factor $s = |\text{metric}_{\text{base}}|$ (or $1.0$ when the base metric is below $0.01$) keeps the adjustment proportional to the magnitude of the base metric. The tuning reward $r_{\text{hpo}}$ depends on how much budget remains: weak tuning is penalized throughout the search, but strong tuning is rewarded only in its later stages. (\Cref{fig:hpo-control-loop}).

The \textbf{combined intervention} applies both the prompt-level directive and the control-loop mechanism together. Full prompts for the directive and the LLM scorer are given in \Cref{sec:hyperparam_appendix}.

\subsection{Thompson sampling and backtracking}

\begin{figure}[H]
  \centering
  \includegraphics[width=0.75\linewidth]{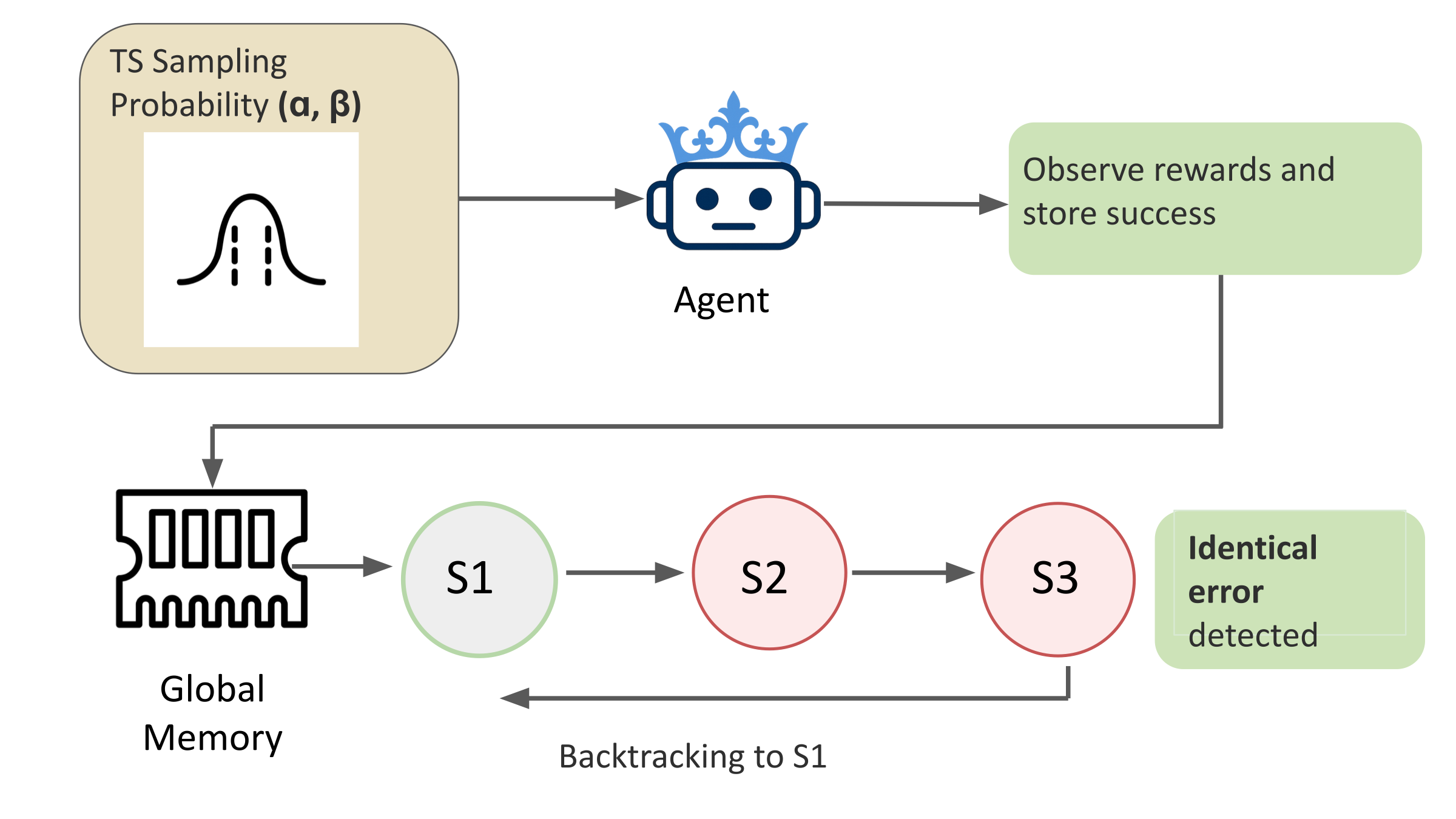}
  \caption{\textbf{Thompson Sampling with backtracking reallocates budget away from repeatedly failing branches.} Each sibling node carries a Beta distribution over its quality, stored in global memory and updated from observed rewards after every execution. When the agent detects an identical error recurring down a path (S1 → S2 → S3), it backtracks to the branch point where the error first appeared (S1) and re-samples among the siblings, expanding the node with the highest draw rather than choosing at random. Budget that would otherwise be spent re-deriving the same failure is redirected toward more promising candidates.}
  \label{fig:ts_back}
\end{figure}

Each of the agents we study maintains a tree of code variants, where every node is a distinct solution attempt with an associated validation metric. When the agent hits the same error repeatedly, we backtrack to the branch point where that error first appeared and reconsider its sibling nodes, rather than continuing to burn budget down a failing path.  Instead of choosing at random as is default with current agents, we use Thompson Sampling (\Cref{fig:ts_back}): each sibling $i$ carries a $\text{Beta}(\alpha_i, \beta_i)$ distribution over its quality, initialized from a uniform $\text{Beta}(1,1)$ prior. At each selection step, we draw a sample $\theta_i \sim \text{Beta}(\alpha_i, \beta_i)$ from every candidate and expand the one with the highest draw, $s^* = \arg\max_i \theta_i$. After the chosen node executes, we score it with a normalized reward $r \in [0,1]$ (0 for buggy nodes, linearly scaled by validation performance otherwise) and update its distribution:

\begin{gather}
  \alpha_{\text{new}} = \alpha_{\text{old}} + r \\
  \beta_{\text{new}} = \beta_{\text{old}} + (1-r)
\end{gather}

This procedure balances exploration (nodes with high uncertainty have wide distributions, giving them higher chances to be sampled) and exploitation (nodes with consistently good performance accumulate higher $\alpha$ values, shifting their distributions rightward), allowing the agent to learn which code branches are more promising with minimal sample complexity.

\section{Experimental setup and results}

\subsection{Experimental setup}
We evaluate our approaches on nine tabular prediction tasks spanning both classification and regression, drawn from MLE-bench and additional Kaggle competitions. Tasks were selected to satisfy three criteria: (i) their release postdates the knowledge cutoff of the underlying LLM, minimizing data leakage; (ii) they are of moderate dataset size to ensure tractable experimentation; and (iii) they collectively cover diverse evaluation metrics. The nine tasks are Cirrhosis Outcome Prediction, GNSS Classification, Spaceship Titanic, Wine Quality, and Playground Series S5E3, S5E6, S5E7, S5E8, and S5E12, with links provided in \Cref{tab:competitions} from \Cref{sec:competition-links}.
We primarily evaluate two agent frameworks, AIDE and ML-Master, both powered by GPT-5-mini. Performance is measured using the official MLE-bench grading scripts, which compute task-specific metrics consistent with each competition (e.g., accuracy, AUC, RMSE) on a held-out test set. All scores are averaged over 10 independent runs with different random seeds; higher scores indicate better performance on all tasks except Cirrhosis Outcome Prediction, where lower is better. Following the MLE-bench medal system, a run earns a gold medal if its score places in the top 10\% of the human leaderboard for that competition. All runs are executed with a fixed compute budget of 2 hours on 22 CPU cores.

\textbf{Note on API models.} All experiments use a single backbone,
GPT-5-mini, because evaluations are prohibitively expensive: the full study (three agents, intervention and baseline conditions, nine competitions, ten seeds each) required over a thousand two-hour runs, and repeating it on a frontier model such as GPT-5.5 or Claude Opus 4.8, which cost roughly ten times more per token, would cost tens of thousands of dollars.

\subsection{Context-aware debugging}
\label{sec:context-debugging}

The debug consultant design produces large and consistent gains for both agents (\Cref{tab:aide-comparison}). For AIDE, the debug consultant nearly doubles the gold-medal count, from 22 to 38, and eliminates all 17 of the baseline's failed runs, raising the valid-submission rate from 81\% to 100\%; ML-Master improves comparably, from 18 to 29 golds, and both agents gain on six of nine competitions. The improvement is largest precisely where context isolation had been most costly: on S5E3 (AIDE) and GNSS (ML-Master), where the baseline earns no medals at all, the consultant recovers a perfect 10/10. It also reaches a working solution far sooner---the median number of steps to a first valid submission drops from 6 to 0, as the consultant supplies the environment's constraints before the agent writes its first line of code.

\begin{table}[t]
\centering
\small
\setlength{\tabcolsep}{3pt}
\begin{minipage}[t]{0.49\linewidth}
\centering
\textbf{(a) AIDE}\\[4pt]
\begin{tabular}{@{}lcccc@{}}
\toprule
Comp. & Treatment & $n$ & Baseline & $n$ \\
\midrule
Cirrhosis$^{\dagger}$ & $\mathbf{0.384 \pm 0.008}$ & 10 & $0.394 \pm 0.035$ & 9 \\
GNSS             & $\mathbf{0.967 \pm 0.002}$ & 10 & $\mathbf{0.967 \pm 0.002}$ & 8 \\
Spaceship & $\mathbf{0.806 \pm 0.016}$ & 10 & $0.790 \pm 0.067$ & 9 \\
Wine    & $\mathbf{0.407 \pm 0.034}$ & 10 & $0.375 \pm 0.059$ & 8 \\
S5E3    & $\mathbf{0.954 \pm 0.015}$ & 10 & $0.888 \pm 0.010$ & 9 \\
S5E6    & $\mathbf{0.337 \pm 0.005}$ & 10 & $0.333 \pm 0.005$ & 8 \\
S5E7    & $\mathbf{0.969 \pm 0.001}$ & 10 & $0.968 \pm 0.000$ & 9 \\
S5E8             & $0.967 \pm 0.009$ & 10 & $\mathbf{0.969 \pm 0.001}$ & 6 \\
S5E12            & $0.725 \pm 0.005$ & 10 & $\mathbf{0.727 \pm 0.001}$ & 7 \\
\midrule
\textbf{W/L}     & \multicolumn{4}{c}{\textbf{6 : 3}} \\
\textbf{Valid}    & \multicolumn{2}{c}{90/90} & \multicolumn{2}{c}{73/90} \\
\textbf{Golds}    & \multicolumn{2}{c}{38} & \multicolumn{2}{c}{22} \\
\bottomrule
\end{tabular}
\end{minipage}%
\hfill
\begin{minipage}[t]{0.49\linewidth}
\centering
\textbf{(b) ML-Master}\label{tab:mlmaster-comparison}\\[4pt]
\begin{tabular}{@{}lcccc@{}}
\toprule
Comp. & Treatment & $n$ & Baseline & $n$ \\
\midrule
Cirrhosis$^{\dagger}$ & $\mathbf{0.380 \pm 0.002}$ & 10 & $0.386 \pm 0.006$ & 10 \\
GNSS    & $\mathbf{0.953 \pm 0.012}$ & 10 & $0.813 \pm 0.002$ & 10 \\
Spaceship & $\mathbf{0.813 \pm 0.008}$ & 10 & $0.794 \pm 0.033$ & 10 \\
Wine             & $\mathbf{0.412 \pm 0.015}$ & 10 & $\mathbf{0.412 \pm 0.028}$ & 10 \\
S5E3             & $0.851 \pm 0.011$ & 10 & $\mathbf{0.860 \pm 0.005}$ & 10 \\
S5E6    & $\mathbf{0.336 \pm 0.003}$ & 10 & $0.326 \pm 0.006$ & 10 \\
S5E7             & $\mathbf{0.968 \pm 0.001}$ & 10 & $\mathbf{0.968 \pm 0.001}$ & 10 \\
S5E8    & $\mathbf{0.969 \pm 0.000}$ & 10 & $0.966 \pm 0.002$ & 10 \\
S5E12   & $\mathbf{0.728 \pm 0.000}$ & 10 & $0.719 \pm 0.011$ & 10 \\
\midrule
\textbf{W/L}     & \multicolumn{4}{c}{\textbf{6 : 3}} \\
\textbf{Valid}    & \multicolumn{2}{c}{90/90} & \multicolumn{2}{c}{90/90} \\
\textbf{Golds}    & \multicolumn{2}{c}{29} & \multicolumn{2}{c}{18} \\
\bottomrule
\end{tabular}
\end{minipage}
\caption{%
  \textbf{The debug consultant improves gold medal rates on both agents.}
  Treatment vs.\ vanilla baseline on 9 MLE-bench competitions $\times$ 10 seeds.
  Scores are mean $\pm$ std; $n$ = seeds with valid submissions.
  Bold = winner by mean; values tied at the displayed precision are bolded in both columns.
  $^{\dagger}$ = lower is better.
}
\label{tab:aide-comparison}
\end{table}

\paragraph{Mechanism.}
The consultant works by stopping the agent from paying for the same mistake twice. In AIDE's search journals, redundant bug encounters fall from 46\% to 7.8\%, and the fraction of nodes that execute without error rises from 54.7\% to 79.0\%. The compute the baseline spends rediscovering known bugs is instead spent producing working code, and this is what improves final scores: seeds with more valid nodes achieve better held-out results (pooled $r = +0.22$ across 163 seeds). Detailed statistics, including first-attempt fix rates and time to first submission, are in \Cref{sec:appendix-debugging-detailed}; case studies, full generated code, and per-competition correlations are in \Cref{sec:appendix-case-studies,sec:appendix-full-code,sec:appendix-correlation}.

\subsection{Hyperparameter tuning guidance}

\begin{table*}[h!]
  \centering
  {\small\setlength{\tabcolsep}{6pt}%
    \begin{tabular}{@{}l c c c@{}}
      \toprule
      \textbf{Competition} & \textbf{$\Delta$Prompt} & \textbf{$\Delta$Code} & \textbf{$\Delta$P\&C} \\
      \midrule
      Cirrhosis$^{\dagger}$ & $+0.035 \pm 0.0123$ & $+0.028 \pm 0.0118$ & $+0.024 \pm 0.0117$ \\
      GNSS & $-0.036 \pm 0.0021$ & $-0.032 \pm 0.0012$ & $-0.030 \pm 0.0010$ \\
      Spaceship & $\boldsymbol{+0.110} \pm 0.0225$ & $\boldsymbol{+0.105} \pm 0.0226$ & $\boldsymbol{+0.098} \pm 0.0235$ \\
      Wine & $\boldsymbol{+0.100} \pm 0.0219$ & $\boldsymbol{+0.047} \pm 0.0270$ & $\boldsymbol{+0.081} \pm 0.0260$ \\
      S5E3 & $\boldsymbol{+0.014} \pm 0.0052$ & $\boldsymbol{+0.004} \pm 0.0064$ & $\boldsymbol{+0.011} \pm 0.0055$ \\
      S5E6 & $\boldsymbol{+0.067} \pm 0.0020$ & $\boldsymbol{+0.063} \pm 0.0030$ & $\boldsymbol{+0.064} \pm 0.0040$ \\
      S5E7 & $\boldsymbol{+0.097} \pm 0.0002$ & $\boldsymbol{+0.097} \pm 0.0002$ & $\boldsymbol{+0.097} \pm 0.0002$ \\
      S5E8 & $\boldsymbol{+0.384} \pm 0.0017$ & $\boldsymbol{+0.387} \pm 0.0003$ & $\boldsymbol{+0.388} \pm 0.0005$ \\
      S5E12 & $\boldsymbol{+0.218} \pm 0.0005$ & $\boldsymbol{+0.218} \pm 0.0009$ & $\boldsymbol{+0.218} \pm 0.0006$ \\
      \bottomrule
    \end{tabular}%
  }
  \caption{\textbf{Hyperparameter guidance yields large performance gains for AIDE across most competitions.} Intervention effect on graded score for AIDE: $\Delta = \mu_{\mathrm{int}}-\mu_{\mathrm{base}}$ (mean $\pm$ SEM of $\Delta$). P\&C = prompt and code. $^{\dagger}$Lower is better for Cirrhosis.}
  \label{tab:results-aide}
\end{table*}

Adding explicit hyperparameter-tuning guidance to AIDE also produces sizable gains, improving graded scores on 7 of 9 competitions with individual effects as large as $+0.388$ on S5E8 and $+0.218$ on S5E12 (\Cref{tab:results-aide}). The gains are concentrated on tasks where the baseline leaves the most room for improvement (S5E8, S5E12, Spaceship, Wine), confirming that AIDE under-invests in tuning and that a modest amount of structured guidance recovers measurable unrealized performance. On tasks where the baseline is already strong, additional tuning guidance produces little response.

The same control-loop intervention that helps AIDE can degrade ML-Master. This is due to the fact that it pushes ML-Master toward an HPO implementation that crashes. ML-Master's memory then records the crash as buggy without propagating why, resulting in the agent continually retrying variants of the same broken approach. Scaffold interventions therefore do not transfer for free: whether one helps depends on interactions between the different components comprising a scaffold. We analyze this asymmetry in detail in \Cref{sec:hyperparam_appendix}.

\subsection{Thompson sampling with backtracking for stable, more reliable search}

Thompson Sampling (TS) replaces the random sibling selection used by current
agents with a strategy that concentrates exploration on promising branches and
backtracks out of repeatedly failing ones. \textbf{Its primary effect is
stability: in a controlled comparison with all other settings held fixed, TS more than halves the number of null runs at almost no cost to peak performance.} Two controlled
studies below, on AIDE and on MLEvolve \citep{du2026mlevolveselfevolvingframeworkautomated}, attribute this gain to the selection
strategy itself.

We achieve this with a path-structuring algorithm for node selection, augmented by a single
new parameter: \textbf{similar\_error\_backtracking\_threshold}, which lets the agent
backtrack to the node level that triggered the first instance of a repeated error,
reclaiming budget that would otherwise be spent re-deriving the same failure. We also widen
the initial drafting phase, raising the number of initial solution nodes from 5 in the
baseline to 20 for TS, since TS realizes its advantage only when it has a rich pool of
candidates to allocate exploration across. Full hyperparameter settings for the baseline and
TS are given in \Cref{tab:hyperparameter}; headline results are reported in
\Cref{tab:avg_comparison}, with standard deviations and null rates in
\Cref{sec:MCTS_extended} (\Cref{tab:std_comparison,tab:null_rates}).

Since we make multiple changes to the agentic pipeline, including introducing new hyper-parameters, we conduct a controlled study to isolate
the contribution of TS itself. We re-evaluate TS with the number of initial drafts and all
other hyperparameters held fixed across conditions, so any observed gain is attributable to
TS alone. For AIDE, we set the native max-debug-depth parameter equal to our similar-error
backtracking threshold, which renders the latter inactive, and compare two head-to-head
conditions: AIDE with increased drafts but no TS, and the full AIDE+TS configuration
(\Cref{tab:aide_ts_drafts}).

AIDE with 20 initial drafts already improves over the plain setting. On top of
that, TS contributes a distinct and practically important advantage.
\textbf{Holding all other settings identical, AIDE+TS reduces null runs from 33 to 15 of 90 relative to AIDE+more drafts (a 54.5\% reduction), delivering markedly more stable outcomes.} On the competitions most sensitive to
exploration, TS maintains or improves scores even
against the stronger draft-augmented baseline, so the added stability comes at no cost to
peak performance.

To further validate TS independently of any hyperparameter changes, we ran a further experiment with
MLEvolve~\citep{du2026mlevolveselfevolvingframeworkautomated}, one of the strongest open-source agents on MLE-bench. Here the
baseline and TS variants share identical configurations throughout; only the candidate
selection strategy differs. Across nine benchmark competitions, MLEvolve+TS
outperforms baseline MLEvolve on 5 out of 9 competitions (\Cref{tab:mlevolve_ts}) and ties a sixth, confirming
that the edge is attributable to TS rather than to incidental tuning.

\begin{table}[h]
  \centering
  \small
  \begin{tabular}{lccc}
    \toprule
    \textbf{Competition} & \textbf{MLEvolve + TS} & \textbf{MLEvolve Baseline} & \textbf{Winner} \\
    \midrule
    Cirrhosis$^{\dagger}$ & 0.390 $\pm$ 0.003 & \textbf{0.389 $\pm$ 0.004} & Baseline \\
    GNSS & \textbf{0.953 $\pm$ 0.005} & 0.928 $\pm$ 0.011 & TS \\
    Spaceship & \textbf{0.815 $\pm$ 0.002} & 0.814 $\pm$ 0.003 & TS \\
    Wine & 0.406 $\pm$ 0.012 & \textbf{0.425 $\pm$ 0.002} & Baseline \\
    S5E3 & \textbf{0.902 $\pm$ 0.001} & 0.899 $\pm$ 0.001 & TS \\
    S5E6 & \textbf{0.313 $\pm$ 0.007} & 0.311 $\pm$ 0.008 & TS \\
    S5E7 & \textbf{0.968 $\pm$ 0.000} & \textbf{0.968 $\pm$ 0.000} & Tie \\
    S5E8 & 0.963 $\pm$ 0.003 & \textbf{0.964 $\pm$ 0.002} & Baseline \\
    S5E12 & \textbf{0.726 $\pm$ 0.001} & 0.711 $\pm$ 0.004 & TS \\
    \bottomrule
  \end{tabular}
  \caption{\textbf{Swapping in Thompson Sampling alone delivers MLEvolve's largest gains.} With all other settings held fixed, TS wins 5 of 9
    competitions and ties a sixth (S5E7). Its two biggest-margin results, GNSS
    ($+2.5\%$) and S5E12 ($+1.5\%$), are well outside SEM, while most remaining
    differences in either direction fall within SEM; the one substantial exception is
    Wine, where the baseline wins by $+1.9\%$. Bold indicates the winning method per competition; values tied
    at the displayed precision are bolded in both columns. $^{\dagger}$Lower is
  better.}
  \label{tab:mlevolve_ts}
\end{table}

Two mechanisms drive these gains, and they reinforce each other. A wider initial
draft pool gives the agent more candidates to work with, and Thompson Sampling
allocates exploration across them far more effectively than random selection, while
backtracking pulls the agent out of repeatedly failing paths and returns that
budget to promising ones. The AIDE and MLEvolve studies let us see each mechanism on
its own: the draft increase helps by itself, and TS adds a further gain on top,
holding everything else fixed. Together they explain why the full system is both
more stable and stronger than either piece alone.

\subsection{Diagnostic: current agents do not meaningfully act on exploratory data analysis}

Additionally, as a diagnostic, we examine whether agent-based systems meaningfully adhere to one of the canonical stages of the machine learning pipeline: exploratory data analysis (EDA). We observe that most agents operate using a three-stage structure, namely \textit{draft()}, \textit{improve()}, and \textit{debug()}. A recurring pattern across agents is the explicit instruction in all three phases to avoid EDA. We hypothesize that, even if this restriction were lifted, the agents would not effectively utilize insights derived from EDA. To test this hypothesis, we inject the results of a deliberately misleading and low-fidelity exploratory data analysis directly into the agent’s context window. Theoretically, if the agent incorporates this information, such adversarial signals would adversely influence its downstream decisions, for example feature selection. We would therefore expect degraded performance metrics as a consequence of these adversely impacted choices.

To test whether agents incorporate exploratory data analysis (EDA), we injected the results of a controlled, erroneous EDA into the context windows of AIDE and ML-Master and compared their performance against EDA-free baselines. Across all tasks, the performance differences induced by EDA injection were inconsistent and statistically insignificant. Using an LLM-as-a-judge framework (gpt-5-2025-8-07), we further found that agents never conducted EDA on their own in baseline runs and rarely engaged with the injected EDA: in AIDE, the agent acknowledged the malicious EDA in only 21\% of cases and let it affect feature selection in just 5\%. Together, these results indicate that existing agents do not meaningfully act upon or integrate EDA into downstream modeling decisions, suggesting an avenue for improving future agents.  Extended details including the injection format, example messages, and the full evaluation prompts are provided in  \Cref{sec:adversarial_eda_extended}.

\section{Related works}
\label{related_works}

\textbf{AutoML outside of LLM agents.}
AutoML systems can automate algorithm selection and hyperparameter tuning.  For example, \textbf{Auto-sklearn} \citep{feurer2022autosklearn} and \textbf{TPOT} \citep{olson2016tpot} leverage Bayesian optimization and ensemble construction. Approaches like \textbf{FLAML} \citep{wang2021flaml} and \textbf{TabPFN} \citep{hollmann2023tabpfn} focus on low-computational-cost optimization or in-context learning for tabular data, respectively. However, these systems are rigid compared to LLM agents that can implement any algorithm in principle, and they often fail to outperform simple baselines in low-data regimes \citep{knauer2024pmlbmini}. Unlike these fixed-pipeline approaches, we employ LLM-driven agents to dynamically reason about data semantics and debug failures in real time.

\textbf{Autoresearch \& agentic data science.}
Recent agents for machine learning engineering have shifted from linear code
generation to sophisticated tree-search methodologies \citep{chan2024mlebench,
wang2023survey}. A broader line of work pushes toward autoresearch, automating larger
portions of the research loop: DataVoyager \citep{pmlr-v235-majumder24a} uses a
role-based multi-agent architecture to explore and verify hypotheses from a dataset,
DiscoveryBench \citep{majumder2025discoverybench} benchmarks agents on this discovery
task rather than on competition-style modeling, and systems like The AI Scientist
\citep{lu2024aiscientistfullyautomated} attempt the full pipeline from ideation to
writeup. The frameworks we build on instead focus on the modeling stage, optimizing
predictive performance through tree search: \textbf{AIDE} \citep{jiang2025aide},
\textbf{ML-Master} \citep{liu2025mlmaster}, and \textbf{R\&D-Agent}
\citep{yang2025rdagent} navigate complex coding tasks via iterative refinement and
multi-agent parallelization. While effective at exploration, these methods lack
explicit mechanisms for addressing the localization of debugging knowledge, often
leading to repetitive errors in the search tree, a gap we address via a debug
consultant that enables adaptive learning of the execution environment.

\textbf{Self-correction \& context engineering.}
LLMs have demonstrated the ability to ``self-debug'' code via iterative generation \citep{chen2024teaching, yang2024sweagent}. However, in domain-specific tasks, pre-training priors frequently override runtime feedback, causing persistent ``fix loops.'' To mitigate this, we draw on Agentic Context Engineering (ACE) \citep{zhang2025agentic} to treat context as an evolving playbook. Crucially, unlike open-ended agents like \textit{Voyager} \citep{wang2024voyager} that accumulate success skills, our domain necessitates the systematic accumulation of \textit{failures}. We argue that in reward-sparse environments like tabular debugging, learning what not to do (negative constraints) provides just as valuable a signal as sparse successes.

\section{Discussion}

Our findings reveal fundamental limitations of existing agents.  Some of what an agent learns is local to a branch. Much of it, however, is a global property of the environment or problem setting: which library versions are installed, which API signatures are valid, how long a fold of training takes on the available cores, or the features of the dataset at hand. Global facts are invariant across the tree, and re-deriving them per branch or node is redundant. A global memory that separates the two lets a run behave as a single agent rather than many isolated ones, which reduces redundancy. More importantly, memory may pay off in generating better hypotheses based on lessons learned globally across previous nodes. An agent that can recall what it has already ruled out can condition its next hypothesis on the accumulated failures instead of resampling from an unchanged prior. We expect the value of such memory to grow rather than shrink as autoresearch systems become more ambitious.

Existing agents are also limited in their ability to explore the search space.  Tree search algorithms assume that expanding a node produces a novel candidate, but LLMs often write nearly the same program over and over again. When an agent produces dozens of near-identical programs, the tree is wide only on paper. Better selection will therefore help only so much until agents are designed to propose more varied solutions, which may be why our own selection improvements reduce variance more than they raise peak scores.

Reliability is another limitation of current agents, and perhaps the most overlooked. On a meaningful fraction of runs, agents produce no result at all, and the common
practice of averaging over successful runs hides these failures so the waste goes
uncounted. As autoresearch systems take on longer and more expensive tasks, a run
that quietly produces nothing becomes far more costly than one that produces a
mediocre answer. Making agents dependable will matter as much as making them capable.

The interventions we study are deliberately simple, and each recovers substantial performance from the same underlying model, which implies that current agents operate well below the ceiling their language models already permit. As these systems begin to take on more of the research loop, forming hypotheses, designing experiments, and interpreting results, the cost of an agent that forgets what it has learned, proposes what it has already tried, or fails silently will only rise. Closing that gap will require treating memory, diversity, and reliability as first-class objectives of agent design rather than as incidental properties of the scaffold.

\section*{Acknowledgements}
This project was supported by a research award from the Center for AI and Responsible Financial Innovation at Columbia University, by the Columbia Center for AI Technology, and by the Data Science Institute at Columbia University.

\bibliographystyle{colm2026_conference}
\bibliography{references}

\newpage
\appendix
\crefalias{section}{appendix}
\crefalias{subsection}{appendix}
\onecolumn

\section{Context-aware debugging: detailed analysis}
\label{sec:appendix-debugging-detailed}

This section provides a comprehensive analysis of how the debug consultant accumulates a shared model of the execution environment and propagates it across the search tree.
We show that baseline agents waste compute rediscovering the same library incompatibilities and API mismatches across branches, trace how the consultant's accumulated knowledge of the runtime---which library versions are installed, which API signatures are valid, and which code patterns crash---changes agent behavior, and present concrete before/after examples.

\subsection{The environmental blindness problem}
\label{sec:appendix-env-blindness}

Without the debug consultant, each branch must independently discover the execution environment's constraints---which library versions are installed, which API parameters have been deprecated, which code patterns are valid in the current container.
The result is massive redundancy: 46.0\% of baseline nodes waste compute re-encountering bugs that have already been seen within the same seed, compared to only 7.8\% under the consultant.
This redundant re-discovery causes total failure in 17 baseline seeds (zero valid submissions).

\subsection{How adaptive learning changes agent behavior}
\label{sec:appendix-online-learning}

\begin{figure}[ht]
  \centering
  \includegraphics[width=\linewidth]{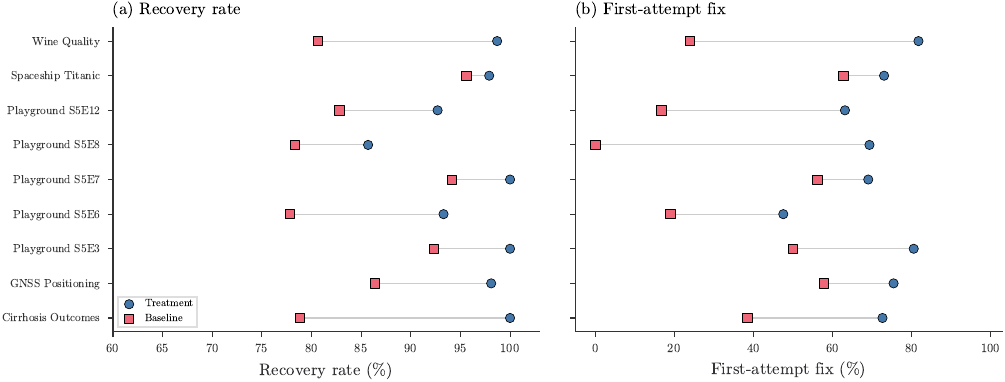}
  \caption{\textbf{The consultant raises first-attempt fix rate from 41.4\% to 72.4\%.} Bug recovery rate (left) and first-attempt fix rate (right) per competition. Constraint injection converts random retries into informed corrections.}
  \label{fig:recovery}
\end{figure}

\Cref{tab:recovery} and \Cref{fig:recovery} show per-competition recovery statistics.
The BANNED list converts what would be random retries into informed corrections: the agent's next attempt is constrained to avoid known-failing patterns, collapsing the search space so it is more likely to succeed.

Treatment recovers from 96.8\% of bug streaks (consecutive buggy nodes), with 72.4\% fixed on the first attempt (average 1.40 attempts).
Baseline recovers from only 86.2\%, with 41.4\% first-attempt fixes and an average of 6.43 attempts.
On the hardest competitions (S5E6, S5E8), baseline first-attempt fix rates are 19.0\% and 0.0\% respectively, while treatment achieves 47.6\% and 69.4\%.

\begin{table}[ht]
\centering
\small
\begin{tabular}{@{}lcccccc@{}}
\toprule
 & \multicolumn{3}{c}{Treatment} & \multicolumn{3}{c}{Baseline} \\
\cmidrule(lr){2-4}\cmidrule(lr){5-7}
Competition & Rec.\% & 1st\% & Att. & Rec.\% & 1st\% & Att. \\
\midrule
Cirrhosis  &  98.2 & 75.9 & 1.31 &  78.8 & 38.5 &  3.69 \\
GNSS       &  98.1 & 75.5 & 1.34 &  86.4 & 57.9 &  3.55 \\
Spaceship  &  97.9 & 73.1 & 1.51 &  95.6 & 62.8 &  2.26 \\
Wine       &  98.7 & 81.8 & 1.22 &  80.6 & 24.0 &  7.00 \\
S5E3       & 100.0 & 80.6 & 1.24 &  92.3 & 50.0 &  3.94 \\
S5E6       &  93.3 & 47.6 & 1.95 &  77.8 & 19.0 & 16.52 \\
S5E7       & 100.0 & 69.1 & 1.38 &  94.1 & 56.2 &  2.12 \\
S5E8       &  85.7 & 69.4 & 1.50 &  78.3 &  0.0 & 19.39 \\
S5E12      &  92.7 & 63.2 & 1.66 &  82.8 & 16.7 & 11.71 \\
\midrule
\textbf{Overall} & \textbf{96.8} & \textbf{72.4} & \textbf{1.40} & 86.2 & 41.4 & 6.43 \\
\bottomrule
\end{tabular}
\caption{
  Bug recovery statistics from AIDE journal analysis.
  Recovery = bug streak followed by $\geq 1$ valid node.
  1st-fix = recovered on the first attempt after the streak.
}
\label{tab:recovery}
\end{table}

\subsection{Synchronization speed}
\label{sec:appendix-sync-speed}

\begin{figure}[ht]
  \centering
  \includegraphics[width=0.85\linewidth]{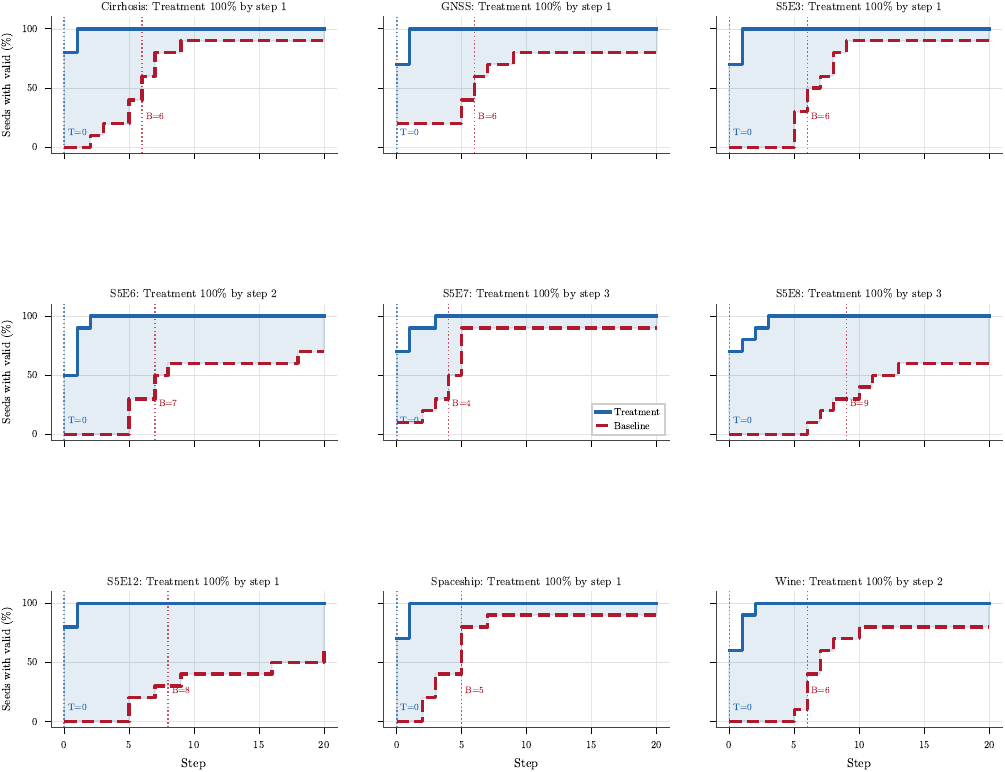}
  \caption{\textbf{Treatment reaches 100\% submission rate by step~3; mean baseline submission rate remains below 40\% through step~5.} Cumulative submission rate by exploration step for treatment vs.\ baseline across all 9 competitions (AIDE, 10 seeds each).}
  \label{fig:submission-rate-v2}
\end{figure}

\Cref{tab:submission-rate} and \Cref{fig:submission-rate-v2} show how quickly the consultant learns the environment.
Treatment seeds produce their first valid node at step~0.4 on average (median: 0)---the consultant has already learned enough about the runtime by the first step to guide the agent past common pitfalls. Baseline seeds, which must rediscover these constraints independently, require 6.8 steps on average (median: 6).
Treatment reaches 94\% submission by step~1 (100\% by step~3); baseline remains at 0\% through step~3 on five competitions.
Under any fixed compute budget $\leq 3$ steps, the scaffolded agent dominates.

\begin{table*}[t]
\centering
\small
\begin{tabular}{@{}lcccc|cccc@{}}
\toprule
 & \multicolumn{4}{c|}{Treatment (\%)} & \multicolumn{4}{c}{Baseline (\%)} \\
Competition & S0 & S1 & S3 & S5 & S0 & S1 & S3 & S5 \\
\midrule
Cirrhosis  &  80 & 100 & 100 & 100 &   0 &   0 &  20 &  40 \\
GNSS       &  70 & 100 & 100 & 100 &  20 &  20 &  20 &  40 \\
Spaceship  &  70 & 100 & 100 & 100 &   0 &   0 &  40 &  80 \\
Wine       &  60 &  90 & 100 & 100 &   0 &   0 &   0 &  10 \\
S5E3       &  70 & 100 & 100 & 100 &   0 &   0 &   0 &  30 \\
S5E6       &  50 &  90 & 100 & 100 &   0 &   0 &   0 &  30 \\
S5E7       &  70 &  90 & 100 & 100 &  10 &  10 &  30 &  90 \\
S5E8       &  70 &  80 & 100 & 100 &   0 &   0 &   0 &   0 \\
S5E12      &  80 & 100 & 100 & 100 &   0 &   0 &   0 &  20 \\
\midrule
\textbf{Mean} & \textbf{69} & \textbf{94} & \textbf{100} & \textbf{100} & 3 & 3 & 12 & 38 \\
\bottomrule
\end{tabular}
\caption{
  Fraction of seeds with $\geq 1$ valid submission by step $N$ (AIDE, 10 seeds per competition).
  Treatment first-valid step: mean 0.4 (median 0). Baseline: mean 6.8 (median 6).
}
\label{tab:submission-rate}
\end{table*}

\subsection{From environmental synchronization to solution quality}
\label{sec:appendix-sync-to-quality}

\Cref{tab:node-stats} shows that the treatment's overall valid rate is 79.0\% (2{,}889 / 3{,}655 nodes) versus 54.7\% (2{,}464 / 4{,}507) for the baseline.
Treatment generates fewer total nodes but a substantially higher fraction are valid, indicating more efficient use of the compute budget. This is partly because valid nodes consume significantly more runtime than invalid ones: a valid node must execute the full pipeline---including data preprocessing, model training (e.g., gradient-boosted trees), and prediction---while an invalid node crashes early and returns quickly. Fewer but valid nodes therefore represent a much larger share of useful compute.

With the environment solved, the LLM focuses on modeling: ensembles, calibration, and problem reformulation emerge through successive refinement.
Per-competition correlations are reported in \Cref{sec:appendix-correlation}.

\begin{table}[ht]
\centering
\small
\begin{tabular}{@{}lrrrrrrr@{}}
\toprule
 & \multicolumn{3}{c}{Treatment} & \multicolumn{3}{c}{Baseline} & \\
\cmidrule(lr){2-4}\cmidrule(lr){5-7}
Comp. & Tot. & Val. & \% & Tot. & Val. & \% & $\Delta$\% \\
\midrule
Cirrhosis  &   500 &   428 & 85.6 &   261 &   150 & 57.5 & $+$28.1 \\
GNSS       &   389 &   317 & 81.5 &   432 &   278 & 64.4 & $+$17.1 \\
Spaceship  &   524 &   380 & 72.5 &   744 &   641 & 86.2 & $-$13.7 \\
Wine       &   530 &   435 & 82.1 &   432 &   234 & 54.2 & $+$27.9 \\
S5E3       &   557 &   474 & 85.1 &   577 &   370 & 64.1 & $+$21.0 \\
S5E6       &   198 &   111 & 56.1 &   434 &    38 &  8.8 & $+$47.3 \\
S5E7       &   597 &   521 & 87.3 &   725 &   651 & 89.8 & $-$2.5  \\
S5E8       &   169 &   100 & 59.2 &   429 &    37 &  8.6 & $+$50.6 \\
S5E12      &   191 &   123 & 64.4 &   473 &    65 & 13.7 & $+$50.7 \\
\midrule
\textbf{Total} & 3{,}655 & 2{,}889 & \textbf{79.0} & 4{,}507 & 2{,}464 & 54.7 & $+$24.3 \\
\bottomrule
\end{tabular}
\caption{
  Node counts from AIDE journal analysis (9 comps $\times$ 10 seeds).
  Treatment generates fewer total nodes but a higher fraction are valid.
}
\label{tab:node-stats}
\end{table}

\section{Extended experimental results}

\subsection{Per-seed OOS scores: AIDE}
\label{sec:appendix-aide-seeds}

\Cref{tab:aide-per-seed} reports the graded out-of-sample score for every AIDE seed (9 competitions $\times$ 10 seeds).
Treatment achieves valid submissions on all 90 seeds; baseline has 17 null runs (``---'').
For Cirrhosis (Log Loss), lower is better; for all others, higher is better.

\begin{table*}[t]
\centering
\scriptsize
\setlength{\tabcolsep}{4pt}
\begin{tabular}{@{}llcccccccccccc@{}}
\toprule
Competition & Cond. & S0 & S1 & S2 & S3 & S4 & S5 & S6 & S7 & S8 & S9 & Mean & Std \\
\midrule
Cirrhosis$^{\dagger}$ & T & 0.3839 & 0.3849 & \textbf{0.3802} & 0.4004 & \textbf{0.3780} & \textbf{0.3767} & \textbf{0.3786} & \textbf{0.3802} & 0.3985 & 0.3803 & 0.384 & 0.008 \\
 & B & \textbf{0.3786} & \textbf{0.3826} & 0.3811 & \textbf{0.3799} & 0.3790 & 0.3847 & 0.3827 & 0.4868 & \textbf{0.3870} & --- & 0.394 & 0.035 \\
\midrule
GNSS & T & \textbf{0.9668} & \textbf{0.9676} & 0.9673 & \textbf{0.9687} & \textbf{0.9684} & 0.9640 & \textbf{0.9672} & 0.9634 & 0.9675 & 0.9649 & 0.967 & 0.002 \\
 & B & 0.9637 & 0.9673 & \textbf{0.9680} & 0.9679 & 0.9668 & \textbf{0.9687} & 0.9627 & --- & \textbf{0.9690} & --- & 0.967 & 0.002 \\
\midrule
S5E3 & T & 0.9590 & \textbf{0.9572} & \textbf{0.9566} & \textbf{0.9634} & \textbf{0.9583} & \textbf{0.9104} & \textbf{0.9515} & \textbf{0.9606} & \textbf{0.9614} & \textbf{0.9570} & 0.954 & 0.015 \\
 & B & --- & 0.8826 & 0.8920 & 0.8771 & 0.8960 & 0.9016 & 0.8748 & 0.8782 & 0.8884 & 0.8966 & 0.887 & 0.010 \\
\midrule
S5E6 & T & \textbf{0.3321} & 0.3328 & \textbf{0.3439} & \textbf{0.3385} & 0.3399 & 0.3405 & 0.3299 & \textbf{0.3407} & 0.3316 & \textbf{0.3413} & 0.337 & 0.005 \\
 & B & 0.3275 & \textbf{0.3387} & 0.3303 & 0.3275 & --- & --- & \textbf{0.3301} & 0.3372 & \textbf{0.3372} & 0.3324 & 0.333 & 0.005 \\
\midrule
S5E7 & T & \textbf{0.9682} & \textbf{0.9692} & \textbf{0.9682} & \textbf{0.9703} & 0.9698 & \textbf{0.9687} & \textbf{0.9698} & \textbf{0.9687} & \textbf{0.9698} & \textbf{0.9703} & 0.969 & 0.001 \\
 & B & \textbf{0.9682} & 0.9676 & \textbf{0.9682} & 0.9676 & --- & 0.9682 & 0.9676 & 0.9682 & 0.9676 & 0.9676 & 0.968 & 0.000 \\
\midrule
S5E8 & T & \textbf{0.9702} & 0.9700 & 0.9693 & 0.9413 & 0.9684 & 0.9692 & 0.9699 & \textbf{0.9696} & 0.9691 & \textbf{0.9698} & 0.967 & 0.009 \\
 & B & 0.9684 & --- & --- & \textbf{0.9697} & \textbf{0.9696} & \textbf{0.9695} & --- & 0.9688 & --- & 0.9691 & 0.969 & 0.001 \\
\midrule
S5E12 & T & 0.7264 & 0.7110 & 0.7262 & 0.7259 & \textbf{0.7264} & 0.7255 & 0.7258 & \textbf{0.7257} & 0.7264 & 0.7268 & 0.725 & 0.005 \\
 & B & \textbf{0.7271} & \textbf{0.7281} & \textbf{0.7271} & \textbf{0.7275} & 0.7257 & \textbf{0.7281} & --- & 0.7256 & --- & --- & 0.727 & 0.001 \\
\midrule
Spaceship & T & \textbf{0.8172} & 0.8138 & 0.8081 & 0.8149 & 0.7920 & \textbf{0.8172} & 0.7897 & 0.8207 & \textbf{0.8149} & 0.7736 & 0.806 & 0.016 \\
 & B & 0.8000 & \textbf{0.8149} & \textbf{0.8218} & \textbf{0.8172} & \textbf{0.8241} & 0.6184 & --- & \textbf{0.8299} & 0.7667 & \textbf{0.8207} & 0.790 & 0.067 \\
\midrule
Wine & T & \textbf{0.4281} & \textbf{0.3980} & \textbf{0.4294} & 0.3321 & \textbf{0.4031} & \textbf{0.4284} & 0.4121 & \textbf{0.4419} & 0.3664 & 0.4315 & 0.407 & 0.034 \\
 & B & 0.3256 & 0.2990 & 0.4249 & \textbf{0.4264} & 0.3728 & 0.3016 & \textbf{0.4181} & 0.4329 & --- & --- & 0.375 & 0.058 \\
\bottomrule
\end{tabular}
\caption{AIDE Per-Seed OOS Scores. Bold = per-seed winner; values tied at the displayed precision are bolded in both rows. --- = null (no valid submission). $^{\dagger}$ = lower is better.}
\label{tab:aide-per-seed}
\end{table*}

\subsection{Per-seed OOS scores: ML-Master}
\label{sec:appendix-mlmaster-seeds}

\Cref{tab:mlmaster-per-seed} reports the graded OOS score for every ML-Master seed.
Both treatment and baseline achieve valid submissions on all 90 seeds.

\begin{table*}[t]
\centering
\scriptsize
\setlength{\tabcolsep}{4pt}
\begin{tabular}{@{}llcccccccccccc@{}}
\toprule
Competition & Cond. & S0 & S1 & S2 & S3 & S4 & S5 & S6 & S7 & S8 & S9 & Mean & Std \\
\midrule
Cirrhosis$^{\dagger}$ & T & \textbf{0.3838} & 0.3795 & \textbf{0.3822} & \textbf{0.3775} & \textbf{0.3802} & 0.3813 & 0.3837 & \textbf{0.3795} & \textbf{0.3813} & \textbf{0.3756} & 0.380 & 0.002 \\
 & B & 0.3856 & \textbf{0.3789} & 0.3882 & 0.3789 & 0.3890 & \textbf{0.3794} & \textbf{0.3821} & 0.3877 & 0.3881 & 0.4000 & 0.386 & 0.006 \\
\midrule
GNSS & T & \textbf{0.9521} & \textbf{0.9617} & \textbf{0.9480} & \textbf{0.9494} & \textbf{0.9576} & \textbf{0.9644} & \textbf{0.9549} & \textbf{0.9617} & \textbf{0.9207} & \textbf{0.9590} & 0.953 & 0.012 \\
 & B & 0.8124 & 0.8148 & 0.8130 & 0.8128 & 0.8140 & 0.8099 & 0.8112 & 0.8128 & 0.8145 & 0.8150 & 0.813 & 0.002 \\
\midrule
S5E3 & T & \textbf{0.8585} & 0.8402 & 0.8539 & 0.8402 & \textbf{0.8721} & 0.8448 & 0.8448 & \textbf{0.8676} & 0.8402 & 0.8493 & 0.851 & 0.011 \\
 & B & \textbf{0.8585} & \textbf{0.8630} & \textbf{0.8585} & \textbf{0.8585} & 0.8676 & \textbf{0.8539} & \textbf{0.8539} & 0.8585 & \textbf{0.8585} & \textbf{0.8676} & 0.860 & 0.005 \\
\midrule
S5E6 & T & \textbf{0.3421} & \textbf{0.3338} & \textbf{0.3343} & \textbf{0.3406} & \textbf{0.3349} & \textbf{0.3358} & \textbf{0.3358} & \textbf{0.3345} & \textbf{0.3309} & 0.3347 & 0.336 & 0.003 \\
 & B & 0.3238 & 0.3297 & 0.3251 & 0.3342 & 0.3256 & 0.3135 & 0.3213 & 0.3243 & 0.3296 & \textbf{0.3352} & 0.326 & 0.006 \\
\midrule
S5E7 & T & \textbf{0.9687} & \textbf{0.9682} & 0.9676 & \textbf{0.9682} & 0.9655 & 0.9644 & \textbf{0.9687} & \textbf{0.9692} & \textbf{0.9682} & 0.9682 & 0.968 & 0.001 \\
 & B & 0.9671 & \textbf{0.9682} & \textbf{0.9682} & \textbf{0.9682} & \textbf{0.9676} & \textbf{0.9660} & 0.9676 & 0.9682 & \textbf{0.9682} & \textbf{0.9687} & 0.968 & 0.001 \\
\midrule
S5E8 & T & \textbf{0.9686} & \textbf{0.9688} & \textbf{0.9686} & 0.9682 & \textbf{0.9693} & \textbf{0.9689} & \textbf{0.9692} & \textbf{0.9697} & \textbf{0.9690} & \textbf{0.9689} & 0.969 & 0.000 \\
 & B & 0.9669 & 0.9638 & 0.9653 & \textbf{0.9683} & 0.9654 & 0.9640 & 0.9650 & 0.9651 & 0.9680 & 0.9683 & 0.966 & 0.002 \\
\midrule
S5E12 & T & \textbf{0.7277} & \textbf{0.7276} & \textbf{0.7278} & \textbf{0.7273} & \textbf{0.7275} & \textbf{0.7274} & \textbf{0.7277} & \textbf{0.7276} & \textbf{0.7279} & \textbf{0.7271} & 0.728 & 0.000 \\
 & B & 0.7258 & 0.7265 & 0.6877 & 0.7258 & 0.7198 & 0.7179 & 0.7180 & 0.7188 & 0.7226 & 0.7258 & 0.719 & 0.011 \\
\midrule
Spaceship & T & \textbf{0.8184} & \textbf{0.8149} & 0.8126 & \textbf{0.8138} & \textbf{0.8207} & \textbf{0.8207} & \textbf{0.8184} & 0.7977 & \textbf{0.8172} & 0.7988 & 0.813 & 0.008 \\
 & B & 0.7287 & 0.7782 & \textbf{0.8138} & 0.8115 & 0.8092 & 0.7356 & 0.8092 & \textbf{0.8161} & \textbf{0.8172} & \textbf{0.8184} & 0.794 & 0.033 \\
\midrule
Wine & T & 0.4011 & 0.4165 & \textbf{0.4181} & 0.3753 & \textbf{0.4243} & 0.4028 & 0.4143 & \textbf{0.4254} & 0.4221 & \textbf{0.4227} & 0.412 & 0.015 \\
 & B & \textbf{0.4318} & \textbf{0.4245} & 0.3308 & \textbf{0.4240} & 0.4240 & \textbf{0.4103} & \textbf{0.4158} & 0.4176 & \textbf{0.4227} & \textbf{0.4227} & 0.412 & 0.028 \\
\bottomrule
\end{tabular}
\caption{\textbf{ML-Master treatment wins the majority of per-seed comparisons despite a fully-valid baseline.} Per-seed OOS scores for 9 competitions $\times$ 10 seeds. Bold = per-seed winner; values tied at the displayed precision are bolded in both rows. $^{\dagger}$ = lower is better.}
\label{tab:mlmaster-per-seed}
\end{table*}

\subsection{Case studies: how valid nodes become better solutions}
\label{sec:appendix-case-studies}

The two case studies below illustrate the same causal chain: the debug consultant removes a persistent API bug $\to$ the agent iterates freely $\to$ it builds qualitatively more sophisticated solutions.
Both cases share a common pattern: the baseline LLM has the same modeling knowledge as the treatment, but it never gets to use it because its compute budget is consumed by redundant bug encounters.

\subsubsection{Case study 1: S5E3 seed 6 (\texorpdfstring{$\Delta = +0.077$}{Delta = +0.077} AUC)}
\label{sec:appendix-case-s5e3}

\begin{figure}[H]
  \centering
  \includegraphics[width=\linewidth]{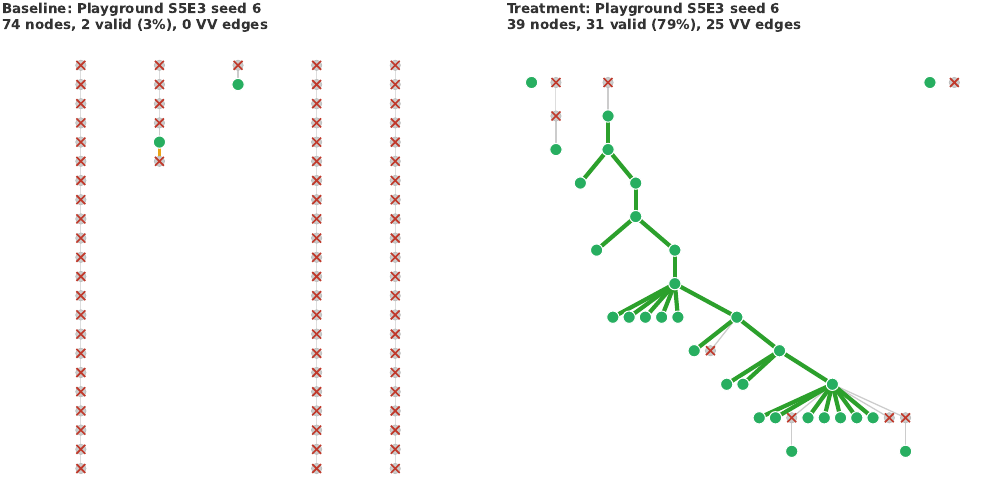}
  \caption{Search tree comparison for S5E3 seed~6. Left: baseline (74 nodes, 2 valid, 2.7\%). Right: treatment (39 nodes, 31 valid, 79.5\%). Red = buggy; green = valid. The baseline loops on the same \texttt{early\_stopping\_rounds} TypeError for its entire budget.}
  \label{fig:paired-tree-s5e3}
\end{figure}

\Cref{fig:paired-tree-s5e3} compares the search trees for S5E3 (binary classification, rainfall prediction), seed~6.

\paragraph{Baseline} (left, 74 nodes, 2 valid, 2.7\%):
The agent's first code draft calls \texttt{lgb.train()} with the deprecated \texttt{early\_stopping\_rounds} parameter.
LightGBM raises a TypeError; the agent's try/except handler catches it silently.
Having no mechanism to propagate this failure, the agent generates nearly identical code 72 more times---each node re-encountering the same TypeError.
Only 2 nodes avoid the pattern, producing a \textbf{single-model LGBMClassifier} with basic feature engineering (4 domain interactions, no calibration, no ensemble).
OOS score: 0.8748.

\paragraph{Treatment} (right, 39 nodes, 31 valid, 79.5\%):
The debug consultant records the \texttt{early\_stopping\_rounds} TypeError at step~0 and adds it to the BANNED list.
From that point forward, the agent never regenerates this pattern, enabling 31 valid iterations to explore progressively better approaches.
The final solution is a \textbf{3-model ensemble} (XGBoost + 2 LightGBM variants) with Platt-scaling calibration and greedy weight optimization. OOS score: 0.9515 ($\Delta = +0.0767$).
With the debug consultant eliminating redundant failures, the treatment refines its approach through 15$\times$ more hypothesis-testing cycles, converging on a substantially more sophisticated solution.

\subsubsection{Case study 2: Wine Quality (\texorpdfstring{$\Delta = +0.099$}{Delta = +0.099} QWK)}
\label{sec:appendix-case-wine}

\begin{figure}[H]
  \centering
  \includegraphics[width=\linewidth]{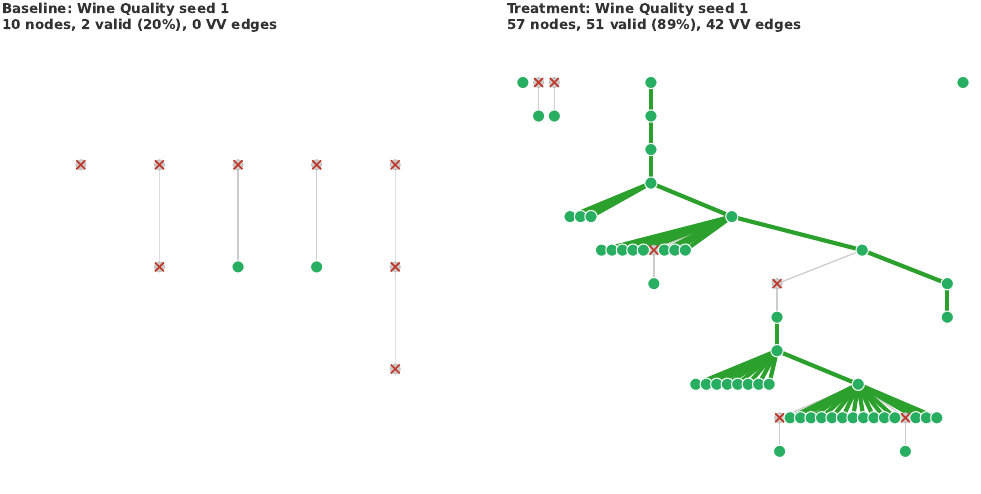}
  \caption{Search tree comparison for Wine seed~1. Left: baseline (10 nodes, 2 valid, 20\%). Right: treatment (57 nodes, 51 valid, 89\%). The baseline exhausts its budget on repeated \texttt{early\_stopping\_rounds} TypeErrors.}
  \label{fig:paired-tree-wine}
\end{figure}

\Cref{fig:paired-tree-wine} compares the search trees for Wine (ordinal classification, 7 quality levels), seed~1.

\paragraph{Baseline} (left, 10 nodes, 2 valid, 20\%):
The agent calls \texttt{LGBMRegressor.fit()} with \texttt{early\_stopping\_rounds} as a keyword argument---deprecated in the installed LightGBM version.
Eight of 10 nodes crash with deprecated-API TypeErrors (7 from \texttt{early\_stopping\_rounds}, 1 from \texttt{verbose}).
The 2 surviving nodes use a single LGBMRegressor with naive rounding of continuous predictions to integer quality labels.
No stacking, no calibration, no threshold optimization.
OOS score: 0.2990.

\paragraph{Treatment} (right, 57 nodes, 51 valid, 89\%):
The consultant's BANNED list prevents the \texttt{early\_stopping\_rounds} TypeError from step~0.
With 51 valid iterations, the agent builds a diverse bagged LightGBM ensemble with stacking (Ridge + Isotonic regression on out-of-fold predictions) and threshold optimization that maps continuous predictions to discrete quality labels, maximizing QWK directly.
OOS score: 0.3980 ($\Delta = +0.099$).

\paragraph{The modeling insight gap.}
Both baseline and treatment use a regression formulation.
The crucial difference is what the agent does with additional valid iterations: it discovers that stacking multiple LightGBM models with isotonic calibration and optimized thresholds produces substantially higher QWK than a single model with naive rounding.
With 51 valid nodes, the treatment has the budget to make this discovery; with 2 valid nodes, the baseline never gets the chance.

\subsubsection{Synthesis: more valid nodes \texorpdfstring{$\to$}{→} better solutions}

Both case studies tell the same story through different competitions and bug types:

\begin{table}[ht]
\centering
\small
\begin{tabular}{@{}lcc@{}}
\toprule
 & S5E3 seed 6 & Wine seed 1 \\
\midrule
Blocking bug & \texttt{early\_stopping\_rounds} & \texttt{early\_stopping\_rounds} \\
Baseline valid nodes & 2 / 74 (2.7\%) & 2 / 10 (20\%) \\
Treatment valid nodes & 31 / 39 (79.5\%) & 51 / 57 (89\%) \\
Baseline approach & Single LGB & LGB regression \\
Treatment approach & 3-model ensemble & Bagged ensemble + threshold opt. \\
OOS $\Delta$ & $+0.077$ & $+0.099$ \\
\bottomrule
\end{tabular}
\caption{Case study summary. The debug consultant's primary effect is enabling iteration.}
\label{tab:case-summary}
\end{table}

We find the same pattern in both cases: adaptive learning allows the LLM to iterate.
A single valid node produces a first working draft.
Dozens of valid nodes allow the agent to discover ensembles, calibration, problem reformulation, and threshold optimization.

\subsection{Full generated code}
\label{sec:appendix-full-code}

Below are the complete, unedited solutions generated by the LLM agent for both case studies.
Each listing is the exact code that was executed and graded.

\subsubsection{S5E3 seed 6 --- Baseline (step~20, OOS = 0.875)}

Note line~88: ``no early stopping due to environment limitations''---the agent is aware of the API bug but cannot resolve it.

\begin{lstlisting}[style=pythonstyle, caption={Baseline best solution: S5E3 seed 6 (141 lines). Single LGBMClassifier with basic feature engineering. No ensemble\, no calibration\, no early stopping.}, label={lst:baseline-s5e3}]
#!/usr/bin/env python3
import os
import warnings
import numpy as np
import pandas as pd
from sklearn.model_selection import StratifiedKFold
from sklearn.metrics import roc_auc_score
from lightgbm import LGBMClassifier

warnings.filterwarnings("ignore")

SEED = 42
N_SPLITS = 5

# Paths
INPUT_DIR = "./input"
SUBMISSION_DIR = "./submission"
WORKING_DIR = "./working"
os.makedirs(SUBMISSION_DIR, exist_ok=True)
os.makedirs(WORKING_DIR, exist_ok=True)

# Load data
train_path = os.path.join(INPUT_DIR, "train.csv")
test_path = os.path.join(INPUT_DIR, "test.csv")

train = pd.read_csv(train_path)
test = pd.read_csv(test_path)


# Feature engineering
def prepare_df(df):
    df = df.copy()
    # day cyclic encoding (assume 365-day cycle)
    if "day" in df.columns:
        df["day_sin"] = np.sin(2 * np.pi * df["day"] / 365.0)
        df["day_cos"] = np.cos(2 * np.pi * df["day"] / 365.0)
    # temperature range
    if {"maxtemp", "mintemp"}.issubset(df.columns):
        df["temp_range"] = df["maxtemp"] - df["mintemp"]
    # dewpoint gap to temp (note column 'temparature' exists in dataset)
    if {"temparature", "dewpoint"}.issubset(df.columns):
        df["temp_minus_dew"] = df["temparature"] - df["dewpoint"]
    # pressure * humidity interaction
    if {"pressure", "humidity"}.issubset(df.columns):
        df["press_hum"] = df["pressure"] * df["humidity"] / 1e3
    # wind * cloud interaction
    if {"windspeed", "cloud"}.issubset(df.columns):
        df["wind_cloud"] = df["windspeed"] * df["cloud"] / 100.0
    return df


train = prepare_df(train)
test = prepare_df(test)

TARGET = "rainfall"
ID_COL = "id"

# Features to use (exclude id, target)
excluded = {ID_COL, TARGET}
features = [c for c in train.columns if c not in excluded]

# Ensure same engineered columns exist in test; if missing add zeros
for f in features:
    if f not in test.columns:
        test[f] = 0.0

X = train[features].reset_index(drop=True)
y = train[TARGET].values
X_test = test[features].reset_index(drop=True)
test_ids = test[ID_COL].values

# Fill NA just in case
X = X.fillna(-999)
X_test = X_test.fillna(-999)

# Prepare CV
skf = StratifiedKFold(n_splits=N_SPLITS, shuffle=True, random_state=SEED)
oof = np.zeros(len(X))
test_preds = np.zeros(len(X_test))

fold_aucs = []

# LGBMClassifier params (sklearn API) - remove early stopping related args
model_params = {
    "objective": "binary",
    "boosting_type": "gbdt",
    "learning_rate": 0.05,
    "n_estimators": 1000,  # train full number of trees (no early stopping in this environment)
    "num_leaves": 31,
    "min_child_samples": 20,
    "subsample": 0.8,
    "colsample_bytree": 0.8,
    "random_state": SEED,
    "n_jobs": -1,
    "verbosity": -1,
}

print(
    "Starting CV training with LGBMClassifier (no early stopping due to environment limitations)..."
)

for fold, (tr_idx, val_idx) in enumerate(skf.split(X, y), 1):
    print(f"\nFold {fold}")
    X_tr, X_val = X.iloc[tr_idx], X.iloc[val_idx]
    y_tr, y_val = y[tr_idx], y[val_idx]

    clf = LGBMClassifier(**model_params)

    # Fit without unsupported kwargs (no eval_set / early_stopping_rounds)
    clf.fit(X_tr, y_tr)

    # Predict validation
    val_pred = clf.predict_proba(X_val)[:, 1]
    oof[val_idx] = val_pred
    fold_auc = roc_auc_score(y_val, val_pred)
    fold_aucs.append(fold_auc)
    print(f"Fold {fold} AUC: {fold_auc:.6f}")

    # Predict test
    test_pred = clf.predict_proba(X_test)[:, 1]
    test_preds += test_pred / N_SPLITS

# Overall OOF AUC
oof_auc = roc_auc_score(y, oof)
print("\nCross-validation results:")
print(f"OOF AUC: {oof_auc:.6f}")
print(f"Mean fold AUC: {np.mean(fold_aucs):.6f}  Std: {np.std(fold_aucs):.6f}")

# Prepare submission
submission = pd.DataFrame({"id": test_ids, "rainfall": test_preds})
submission_path = os.path.join(SUBMISSION_DIR, "submission.csv")
submission.to_csv(submission_path, index=False)
# also save a copy in working for safety
submission.to_csv(os.path.join(WORKING_DIR, "submission.csv"), index=False)

print(f"\nSaved submission to {submission_path}")
print("Submission head:")
print(submission.head())

# Print OOF AUC explicitly as the main evaluation metric
print(f"\nFinal reported OOF ROC AUC: {oof_auc:.6f}")
\end{lstlisting}

\subsubsection{S5E3 seed 6 --- Treatment (step~26, OOS = 0.952)}

The treatment's 31 valid iterations enable it to discover: (1)~a 3-model ensemble (XGBoost + 2 LightGBM variants), (2)~Platt-scaling calibration on each model's OOF predictions, (3)~greedy coordinate-ascent weight optimization in both probability-space and rank-space, and (4)~automatic selection of the better ensemble via calibrated OOF AUC.

\begin{lstlisting}[style=pythonstyle, caption={Treatment best solution: S5E3 seed 6 (437 lines). 3-model ensemble with Platt-scaling calibration\, greedy weight optimization\, and automatic ensemble selection.}, label={lst:treatment-s5e3}]
#!/usr/bin/env python3
import os
import json
from pathlib import Path
import warnings
import numpy as np
import pandas as pd
from sklearn.model_selection import StratifiedKFold, KFold
from sklearn.metrics import roc_auc_score
from sklearn.linear_model import LogisticRegression
import xgboost as xgb
import lightgbm as lgb

warnings.filterwarnings("ignore")

# Config
SEED = 42
N_SPLITS = 5
THREADS = int(os.getenv("AIDE_NUM_THREADS", "22"))
INPUT_DIR = Path("input")
WORKING_DIR = Path("working")
WORKING_DIR.mkdir(parents=True, exist_ok=True)

TRAIN_PATH = INPUT_DIR / "train.csv"
TEST_PATH = INPUT_DIR / "test.csv"
SUBMISSION_PATH = WORKING_DIR / "submission.csv"

# Model hyperparameters (compact, tuned for CPU)
XGB_PARAMS = {
    "objective": "binary:logistic",
    "eval_metric": "auc",
    "verbosity": 0,
    "seed": SEED,
    "eta": 0.03,
    "max_depth": 6,
    "subsample": 0.8,
    "colsample_bytree": 0.8,
    "nthread": THREADS,
    "tree_method": "hist",
}
XGB_NUM_ROUNDS = 1000
XGB_ESR = 50

LGB_PARAMS_A = {
    "objective": "binary",
    "metric": "auc",
    "learning_rate": 0.03,
    "num_leaves": 31,
    "feature_fraction": 0.8,
    "bagging_fraction": 0.8,
    "bagging_freq": 1,
    "min_data_in_leaf": 20,
    "verbose": -1,
    "seed": SEED,
    "num_threads": THREADS,
}
LGB_PARAMS_B = {
    "objective": "binary",
    "metric": "auc",
    "learning_rate": 0.02,
    "num_leaves": 64,
    "feature_fraction": 0.7,
    "bagging_fraction": 0.7,
    "bagging_freq": 1,
    "min_data_in_leaf": 15,
    "min_sum_hessian_in_leaf": 1e-3,
    "verbose": -1,
    "seed": SEED + 1,
    "num_threads": THREADS,
}
LGB_NUM_ROUNDS = 1000
LGB_ESR = 50

# Load data
train = pd.read_csv(TRAIN_PATH)
test = pd.read_csv(TEST_PATH)

TARGET = "rainfall"
IDCOL = "id"

# Feature setup
drop_cols = [IDCOL, TARGET]
features = [c for c in train.columns if c not in drop_cols]

# Cyclical features for 'day' and 'winddirection' if present
if "day" in features:
    period = 365.0
    train["day_sin"] = np.sin(2 * np.pi * train["day"] / period)
    train["day_cos"] = np.cos(2 * np.pi * train["day"] / period)
    test["day_sin"] = np.sin(2 * np.pi * test["day"] / period)
    test["day_cos"] = np.cos(2 * np.pi * test["day"] / period)
    features = [f for f in features if f != "day"] + ["day_sin", "day_cos"]

if "winddirection" in features:
    period = 360.0
    train["winddir_sin"] = np.sin(2 * np.pi * train["winddirection"] / period)
    train["winddir_cos"] = np.cos(2 * np.pi * train["winddirection"] / period)
    test["winddir_sin"] = np.sin(2 * np.pi * test["winddirection"] / period)
    test["winddir_cos"] = np.cos(2 * np.pi * test["winddirection"] / period)
    features = [f for f in features if f != "winddirection"] + [
        "winddir_sin",
        "winddir_cos",
    ]

# Ensure features present in both
features = [f for f in features if f in train.columns and f in test.columns]

# Missing count feature
train["_missing_count"] = train[features].isnull().sum(axis=1)
test["_missing_count"] = test[features].isnull().sum(axis=1)

# Convert object columns to category codes safely
for col in features:
    if train[col].dtype == "object":
        combined = pd.concat([train[col], test[col]], axis=0).astype("category")
        train[col] = combined.iloc[: len(train)].cat.codes
        test[col] = combined.iloc[len(train) :].cat.codes

# Numeric columns and median imputation
train_nums = train[features].select_dtypes(include=[np.number]).columns.tolist()
medians = train[train_nums].median()
train[train_nums] = train[train_nums].fillna(medians)
test[train_nums] = test[train_nums].fillna(medians)

# Row-level stats
train["_row_mean"] = train[train_nums].mean(axis=1)
train["_row_std"] = train[train_nums].std(axis=1).fillna(0.0)
test["_row_mean"] = test[train_nums].mean(axis=1)
test["_row_std"] = test[train_nums].std(axis=1).fillna(0.0)

# Per-column percentile rank aggregated as a row feature.
rank_cols = train_nums.copy()
if len(rank_cols) > 0:
    combined_ranks = []
    for col in rank_cols:
        combined = pd.concat([train[col], test[col]], axis=0)
        ranks = combined.rank(pct=True, method="average")
        combined_ranks.append(ranks.values)
    combined_ranks = np.vstack(combined_ranks).T
    n_train = train.shape[0]
    train_ranks = combined_ranks[:n_train]
    test_ranks = combined_ranks[n_train:]
    train["_row_rank_mean"] = np.nanmean(train_ranks, axis=1)
    test["_row_rank_mean"] = np.nanmean(test_ranks, axis=1)
else:
    train["_row_rank_mean"] = 0.0
    test["_row_rank_mean"] = 0.0

engineered = ["_missing_count", "_row_mean", "_row_std", "_row_rank_mean"]
for f in engineered:
    if f not in features:
        features.append(f)

features = [f for f in features if f in train.columns and f in test.columns]

# Remove constant features
const_feats = [f for f in features if train[f].nunique() <= 1]
if const_feats:
    features = [f for f in features if f not in const_feats]

# Remove near-duplicate features by very high correlation (>0.999)
if len(features) > 1:
    corr_matrix = train[features].corr().abs()
    upper = corr_matrix.where(np.triu(np.ones(corr_matrix.shape), k=1).astype(bool))
    to_drop = [column for column in upper.columns if any(upper[column] > 0.999)]
    if to_drop:
        features = [f for f in features if f not in to_drop]

features = [f for f in features if f in train.columns and f in test.columns]

X = train[features].copy()
X_test = test[features].copy()
y = train[TARGET].astype(int).values
ids_test = test[IDCOL].values

# Cross-validation splitter
min_class_count = pd.Series(y).value_counts().min()
if min_class_count >= N_SPLITS:
    cv = StratifiedKFold(n_splits=N_SPLITS, shuffle=True, random_state=SEED)
else:
    cv = KFold(n_splits=N_SPLITS, shuffle=True, random_state=SEED)

X_values = X.values
X_test_values = X_test.values

# Base models: XGBoost, LGB A, LGB B (single-seed to save time)
model_keys = ["xgb", "lgb_a", "lgb_b"]
oof_preds = {k: np.zeros(X_values.shape[0], dtype=float) for k in model_keys}
test_preds_avg = {k: np.zeros(X_test_values.shape[0], dtype=float) for k in model_keys}

fold_info = []

for fold, (tr_idx, val_idx) in enumerate(cv.split(X_values, y), start=1):
    X_tr, X_val = X_values[tr_idx], X_values[val_idx]
    y_tr, y_val = y[tr_idx], y[val_idx]

    pos = int(y_tr.sum())
    neg = int(y_tr.shape[0] - pos)
    scale_pos_weight = max(1.0, neg / max(1.0, pos))

    # XGBoost
    xgb_params = XGB_PARAMS.copy()
    xgb_params["scale_pos_weight"] = scale_pos_weight
    xgb_params["seed"] = int(SEED)
    dtrain = xgb.DMatrix(X_tr, label=y_tr, feature_names=features)
    dval = xgb.DMatrix(X_val, label=y_val, feature_names=features)
    dtest = xgb.DMatrix(X_test_values, feature_names=features)
    xgb_model = xgb.train(
        xgb_params,
        dtrain,
        num_boost_round=XGB_NUM_ROUNDS,
        evals=[(dtrain, "train"), (dval, "val")],
        early_stopping_rounds=XGB_ESR,
        verbose_eval=False,
    )
    if hasattr(xgb_model, "best_iteration") and xgb_model.best_iteration is not None:
        xgb_rounds = int(xgb_model.best_iteration) + 1
    else:
        xgb_rounds = XGB_NUM_ROUNDS
    val_pred_xgb = xgb_model.predict(dval, iteration_range=(0, xgb_rounds))
    test_pred_xgb = xgb_model.predict(dtest, iteration_range=(0, xgb_rounds))
    oof_preds["xgb"][val_idx] = val_pred_xgb
    test_preds_avg["xgb"] += test_pred_xgb / N_SPLITS

    # LightGBM A
    lgb_params_a = LGB_PARAMS_A.copy()
    lgb_params_a["scale_pos_weight"] = scale_pos_weight
    lgb_params_a["seed"] = int(SEED)
    ltrain = lgb.Dataset(X_tr, label=y_tr, feature_name=features)
    lval = lgb.Dataset(X_val, label=y_val, reference=ltrain, feature_name=features)
    lgb_model_a = lgb.train(
        lgb_params_a,
        ltrain,
        num_boost_round=LGB_NUM_ROUNDS,
        valid_sets=[ltrain, lval],
        valid_names=["train", "val"],
        callbacks=[lgb.early_stopping(stopping_rounds=LGB_ESR)],
    )
    lgb_a_iter = (
        lgb_model_a.best_iteration
        if hasattr(lgb_model_a, "best_iteration")
        else LGB_NUM_ROUNDS
    )
    val_pred_lgb_a = lgb_model_a.predict(X_val, num_iteration=lgb_a_iter)
    test_pred_lgb_a = lgb_model_a.predict(X_test_values, num_iteration=lgb_a_iter)
    oof_preds["lgb_a"][val_idx] = val_pred_lgb_a
    test_preds_avg["lgb_a"] += test_pred_lgb_a / N_SPLITS

    # LightGBM B
    lgb_params_b = LGB_PARAMS_B.copy()
    lgb_params_b["scale_pos_weight"] = scale_pos_weight
    lgb_params_b["seed"] = int(SEED + 1)
    ltrain_b = lgb.Dataset(X_tr, label=y_tr, feature_name=features)
    lval_b = lgb.Dataset(X_val, label=y_val, reference=ltrain_b, feature_name=features)
    lgb_model_b = lgb.train(
        lgb_params_b,
        ltrain_b,
        num_boost_round=LGB_NUM_ROUNDS,
        valid_sets=[ltrain_b, lval_b],
        valid_names=["train", "val"],
        callbacks=[lgb.early_stopping(stopping_rounds=LGB_ESR)],
    )
    lgb_b_iter = (
        lgb_model_b.best_iteration
        if hasattr(lgb_model_b, "best_iteration")
        else LGB_NUM_ROUNDS
    )
    val_pred_lgb_b = lgb_model_b.predict(X_val, num_iteration=lgb_b_iter)
    test_pred_lgb_b = lgb_model_b.predict(X_test_values, num_iteration=lgb_b_iter)
    oof_preds["lgb_b"][val_idx] = val_pred_lgb_b
    test_preds_avg["lgb_b"] += test_pred_lgb_b / N_SPLITS

    fold_auc_vals = tuple(
        float(roc_auc_score(y_val, oof_preds[k][val_idx])) for k in model_keys
    )
    print(
        f"Fold {fold}: "
        + ", ".join(
            [f"{k.upper()} AUC={a:.6f}" for k, a in zip(model_keys, fold_auc_vals)]
        )
    )
    fold_info.append(fold_auc_vals)

# Base OOF AUCs
auc_oof = {k: float(roc_auc_score(y, oof_preds[k])) for k in model_keys}
print("Base OOF AUCs:", auc_oof)

# Calibrate each base with Platt-scaling (LogisticRegression) on OOF
calibrators = {}
calibrated_oof = {}
calibrated_test = {}
for key in model_keys:
    clf = LogisticRegression(solver="lbfgs", max_iter=2000, random_state=SEED)
    preds = oof_preds[key].reshape(-1, 1)
    clf.fit(preds, y)
    calibrated = clf.predict_proba(preds)[:, 1]
    calibrated_oof[key] = calibrated
    calibrated_test[key] = clf.predict_proba(test_preds_avg[key].reshape(-1, 1))[:, 1]
    calibrators[key] = clf
    auc_cal = float(roc_auc_score(y, calibrated))
    print(f"Calibrated OOF AUC for {key}: {auc_cal:.6f}")

# Prepare rank-space OOFs (percentile ranks of calibrated probs)
calibrated_oof_rank = {}
for k in model_keys:
    calibrated_oof_rank[k] = pd.Series(calibrated_oof[k]).rank(pct=True).values


# Greedy coordinate-ascent weight search function
def greedy_optimize(keys, oof_dict, y, init=None, max_iters=200):
    n = len(keys)
    if init is None:
        weights = np.array([1.0 / n] * n, dtype=float)
    else:
        weights = np.array(init, dtype=float)
        if weights.sum() <= 0:
            weights = np.array([1.0 / n] * n, dtype=float)
        else:
            weights = weights / weights.sum()
    best_auc = roc_auc_score(y, sum(weights[i] * oof_dict[keys[i]] for i in range(n)))
    improved = True
    iters = 0
    while improved and iters < max_iters:
        improved = False
        iters += 1
        for i in range(n):
            for delta in [0.1, 0.05, 0.02, 0.01, -0.01, -0.02, -0.05, -0.1]:
                w_new = weights.copy()
                w_new[i] = max(0.0, w_new[i] + delta)
                if w_new.sum() <= 0:
                    continue
                w_new = w_new / w_new.sum()
                ensemble_oof = sum(w_new[j] * oof_dict[keys[j]] for j in range(n))
                auc = roc_auc_score(y, ensemble_oof)
                if auc > best_auc + 1e-9:
                    best_auc = auc
                    weights = w_new
                    improved = True
    return weights, best_auc


keys = model_keys.copy()

# Optimize in probability-space
weights_prob, auc_prob = greedy_optimize(keys, calibrated_oof, y, max_iters=200)
print(
    f"Optimized weights (prob-space): {dict(zip(keys, weights_prob.round(4)))}  OOF AUC: {auc_prob:.6f}"
)

# Optimize in rank-space
weights_rank, auc_rank = greedy_optimize(keys, calibrated_oof_rank, y, max_iters=200)
print(
    f"Optimized weights (rank-space): {dict(zip(keys, weights_rank.round(4)))}  OOF AUC: {auc_rank:.6f}"
)

# Build ensembles
oof_prob_ens = sum(weights_prob[i] * calibrated_oof[keys[i]] for i in range(len(keys)))
test_prob_ens = sum(
    weights_prob[i] * calibrated_test[keys[i]] for i in range(len(keys))
)

oof_rank_ens = sum(
    weights_rank[i] * calibrated_oof_rank[keys[i]] for i in range(len(keys))
)
# convert rank ensemble to pseudo-prob by scaling between 0-1 (already 0-1)
test_rank_parts = {}
for k in keys:
    # get percentile ranks for test calibrated probs
    test_rank_parts[k] = pd.Series(calibrated_test[k]).rank(pct=True).values
test_rank_ens = sum(
    weights_rank[i] * test_rank_parts[keys[i]] for i in range(len(keys))
)

# Evaluate which ensemble is better by OOF AUC
auc_prob_raw = roc_auc_score(y, oof_prob_ens)
auc_rank_raw = roc_auc_score(y, oof_rank_ens)
print(
    f"Raw ensemble OOF AUCs -> prob-space: {auc_prob_raw:.6f}, rank-space: {auc_rank_raw:.6f}"
)

# Final Platt calibration for both ensembles
final_cal_prob = LogisticRegression(solver="lbfgs", max_iter=2000, random_state=SEED)
final_cal_prob.fit(oof_prob_ens.reshape(-1, 1), y)
oof_prob_cal = final_cal_prob.predict_proba(oof_prob_ens.reshape(-1, 1))[:, 1]
test_prob_cal = final_cal_prob.predict_proba(test_prob_ens.reshape(-1, 1))[:, 1]
auc_prob_cal = roc_auc_score(y, oof_prob_cal)
print(f"Final calibrated prob-space ensemble OOF AUC: {auc_prob_cal:.6f}")

final_cal_rank = LogisticRegression(solver="lbfgs", max_iter=2000, random_state=SEED)
final_cal_rank.fit(oof_rank_ens.reshape(-1, 1), y)
oof_rank_cal = final_cal_rank.predict_proba(oof_rank_ens.reshape(-1, 1))[:, 1]
test_rank_cal = final_cal_rank.predict_proba(test_rank_ens.reshape(-1, 1))[:, 1]
auc_rank_cal = roc_auc_score(y, oof_rank_cal)
print(f"Final calibrated rank-space ensemble OOF AUC: {auc_rank_cal:.6f}")

# Choose best
if auc_prob_cal >= auc_rank_cal:
    chosen_name = "prob_space_ensemble"
    oof_final = oof_prob_cal
    test_final = test_prob_cal
    chosen_auc = auc_prob_cal
else:
    chosen_name = "rank_space_ensemble"
    oof_final = oof_rank_cal
    test_final = test_rank_cal
    chosen_auc = auc_rank_cal

print(f"Chosen ensemble: {chosen_name} with OOF AUC: {chosen_auc:.6f}")

# Compute per-fold AUCs for chosen final predictions
fold_scores = []
for fold, (_, val_idx) in enumerate(cv.split(X_values, y), start=1):
    fold_auc = float(roc_auc_score(y[val_idx], oof_final[val_idx]))
    fold_scores.append(fold_auc)
    print(f"Fold {fold} AUC (final): {fold_auc:.6f}")

cv_mean = float(np.mean(fold_scores))
cv_std = float(np.std(fold_scores))
print(f"Final CV mean AUC: {cv_mean:.6f}  std: {cv_std:.6f}")

# Save submission
submission = pd.DataFrame({IDCOL: ids_test, TARGET: test_final})
submission.to_csv(SUBMISSION_PATH, index=False)
print(f"Saved submission to: {SUBMISSION_PATH}")

# Output AIDE metrics line
aide_metrics = {
    "valid": True,
    "lower_is_better": False,
    "cv_mean": cv_mean,
    "cv_std": cv_std,
    "cv_folds": [float(f) for f in fold_scores],
}
print("AIDE_METRICS_JSON=" + json.dumps(aide_metrics))

# Print final metric line
print(f"Final CV mean AUC: {cv_mean:.6f}")
\end{lstlisting}

\subsubsection{Wine Quality Seed 1 --- Baseline (step~7, OOS = 0.299)}

The baseline uses LGBMRegressor with naive rounding---a regression formulation that discards inter-class probability information, fundamentally limiting QWK.

\begin{lstlisting}[style=pythonstyle, caption={Baseline best solution: Wine Quality seed 1. LGBMRegressor with regression formulation and naive rounding.}, label={lst:baseline-wine}]
import os
import numpy as np
import pandas as pd
from sklearn.model_selection import StratifiedKFold
import lightgbm as lgb
from sklearn.metrics import confusion_matrix

# Set random seed
RANDOM_STATE = 42
np.random.seed(RANDOM_STATE)

# Paths
INPUT_DIR = "./input"
TRAIN_PATH = os.path.join(INPUT_DIR, "train.csv")
TEST_PATH = os.path.join(INPUT_DIR, "test.csv")
SUBMISSION_DIR = "./submission"
SUBMISSION_PATH = os.path.join(SUBMISSION_DIR, "submission.csv")

# Read data
train = pd.read_csv(TRAIN_PATH)
test = pd.read_csv(TEST_PATH)

# Features and target
feature_cols = [c for c in train.columns if c not in ("Id", "quality")]
X = train[feature_cols].copy()
y = train["quality"].astype(int).copy()
X_test = test[feature_cols].copy()
test_ids = test["Id"].astype(int).copy()

# Determine label range
label_min = int(y.min())
label_max = int(y.max())
labels_sorted = np.arange(label_min, label_max + 1)


# Quadratic Weighted Kappa implementation
def quadratic_weighted_kappa(y_true, y_pred, min_rating=None, max_rating=None):
    """
    Compute Quadratic Weighted Kappa (QWK)
    """
    if min_rating is None:
        min_rating = min(int(np.min(y_true)), int(np.min(y_pred)))
    if max_rating is None:
        max_rating = max(int(np.max(y_true)), int(np.max(y_pred)))
    y_true = np.array(y_true, dtype=int)
    y_pred = np.array(y_pred, dtype=int)
    num_ratings = int(max_rating - min_rating + 1)
    # Confusion matrix O
    O = np.zeros((num_ratings, num_ratings), dtype=float)
    for a, b in zip(y_true, y_pred):
        O[a - min_rating, b - min_rating] += 1
    # Histogram of ratings
    hist_true = O.sum(axis=1)
    hist_pred = O.sum(axis=0)
    # Expected matrix E
    E = np.outer(hist_true, hist_pred)
    if E.sum() == 0:
        return 1.0
    E = E / E.sum() * O.sum()
    # Weight matrix
    W = np.zeros((num_ratings, num_ratings), dtype=float)
    for i in range(num_ratings):
        for j in range(num_ratings):
            W[i, j] = ((i - j) ** 2) / ((num_ratings - 1) ** 2)
    # QWK
    num = (W * O).sum()
    den = (W * E).sum()
    if den == 0:
        return 1.0
    return 1.0 - num / den


# 5-fold stratified by label
n_splits = 5
skf = StratifiedKFold(n_splits=n_splits, shuffle=True, random_state=RANDOM_STATE)

oof_preds = np.zeros(len(X), dtype=float)
test_preds = np.zeros(len(X_test), dtype=float)
fold_qwks = []

# LightGBM parameters - use scikit-learn API without early_stopping_rounds in fit
lgb_params = {
    "objective": "regression",
    "boosting_type": "gbdt",
    "learning_rate": 0.05,
    "n_estimators": 800,  # fixed number of trees; no early stopping in fit to avoid compatibility issues
    "random_state": RANDOM_STATE,
    "num_leaves": 31,
    "subsample": 0.8,
    "colsample_bytree": 0.8,
    "reg_alpha": 0.0,
    "reg_lambda": 1.0,
    "verbosity": -1,
}

print("Starting 5-fold CV training (no early stopping in fit)...")
for fold, (train_idx, val_idx) in enumerate(skf.split(X, y), 1):
    X_train, X_val = X.iloc[train_idx], X.iloc[val_idx]
    y_train, y_val = y.iloc[train_idx], y.iloc[val_idx]
    model = lgb.LGBMRegressor(**lgb_params)
    # Fit without early_stopping_rounds to avoid the TypeError seen previously
    model.fit(
        X_train,
        y_train,
        eval_set=[(X_val, y_val)],
        eval_metric="rmse",
    )
    # Predict (do not use num_iteration parameter which may not be available)
    val_pred = model.predict(X_val)
    oof_preds[val_idx] = val_pred
    test_pred = model.predict(X_test)
    test_preds += test_pred / n_splits

    # Convert to discrete for QWK
    val_pred_round = np.rint(val_pred).astype(int)
    val_pred_round = np.clip(val_pred_round, label_min, label_max)
    qwk = quadratic_weighted_kappa(
        y_val.values, val_pred_round, min_rating=label_min, max_rating=label_max
    )
    fold_qwks.append(qwk)
    print(f"Fold {fold} QWK: {qwk:.5f}")

# Overall OOF QWK
oof_preds_round = np.rint(oof_preds).astype(int)
oof_preds_round = np.clip(oof_preds_round, label_min, label_max)
oof_qwk = quadratic_weighted_kappa(
    y.values, oof_preds_round, min_rating=label_min, max_rating=label_max
)
print(f"OOF QWK: {oof_qwk:.5f}")
print(f"Mean per-fold QWK: {np.mean(fold_qwks):.5f} (std {np.std(fold_qwks):.5f})")

# Prepare final test predictions
final_test_preds = np.rint(test_preds).astype(int)
final_test_preds = np.clip(final_test_preds, label_min, label_max)

# Create submission dataframe
submission_df = pd.DataFrame({"Id": test_ids.values, "quality": final_test_preds})
# Ensure submission directory exists
os.makedirs(SUBMISSION_DIR, exist_ok=True)
submission_df.to_csv(SUBMISSION_PATH, index=False)

print(f"Saved submission to {SUBMISSION_PATH}")
print("Sample of submission:")
print(submission_df.head(10).to_string(index=False))
\end{lstlisting}

\subsubsection{Wine Quality Seed 1 --- Treatment (step~38, OOS = 0.398)}

With 51 valid iterations, the treatment discovers that a diverse bagged LightGBM ensemble with stacking (Ridge + Isotonic regression on out-of-fold predictions) produces substantially higher QWK than the baseline's single-model regression approach.

\begin{lstlisting}[style=pythonstyle, caption={Treatment best solution: Wine Quality seed 1. Diverse bagged LightGBM ensemble with stacking.}, label={lst:treatment-wine}]
#!/usr/bin/env python3
import os
import json
import math
import numpy as np
import pandas as pd
from itertools import combinations
from sklearn.model_selection import StratifiedKFold, KFold
from sklearn.metrics import cohen_kappa_score
from sklearn.linear_model import Ridge
from sklearn.isotonic import IsotonicRegression
import lightgbm as lgb
import warnings

warnings.filterwarnings("ignore")

# Config
SEED = 42
NUM_FOLDS = 5
LGB_ROUNDS = 2000
EARLY_STOPPING = 100
THREADS = int(os.getenv("AIDE_NUM_THREADS", "22"))
INPUT_DIR = "input"
TRAIN_PATH = os.path.join(INPUT_DIR, "train.csv")
TEST_PATH = os.path.join(INPUT_DIR, "test.csv")
SUBMISSION_PATH = os.path.join("working", "submission.csv")

np.random.seed(SEED)

# Load data
train = pd.read_csv(TRAIN_PATH)
test = pd.read_csv(TEST_PATH)

FEATURES = [c for c in train.columns if c not in ("id", "quality")]
X_orig = train[FEATURES].copy()
X_test_orig = test[FEATURES].copy()
test_ids = test["id"].astype(int).values

y_raw = train["quality"].astype(int).values
unique_classes = np.sort(train["quality"].unique())
n_classes = len(unique_classes)
n_train = len(train)
n_test = len(test)


# Safe feature engineering (no target leakage)
def add_features(df):
    df2 = df.copy()
    for c in df.columns:
        col = df[c]
        if pd.api.types.is_numeric_dtype(col):
            if (col > 0).all():
                df2[c + "_log1p"] = np.log1p(col)
            else:
                df2[c + "_rankpct"] = col.rank(pct=True).astype(float)
            # square
            df2[c + "_sq"] = col.values * col.values
    return df2


X = add_features(X_orig)
X_test = add_features(X_test_orig)

# pairwise products among top-k variance original features (produce train/test separately)
num_feats = [c for c in X_orig.columns if pd.api.types.is_numeric_dtype(X_orig[c])]
variances = [(c, X_orig[c].var()) for c in num_feats]
variances.sort(key=lambda x: x[1], reverse=True)
top_k = min(5, len(variances))
top_features = [c for c, _ in variances[:top_k]]

for a, b in combinations(top_features, 2):
    name = f"{a}_x_{b}"
    X[name] = X_orig[a].values * X_orig[b].values
    X_test[name] = X_test_orig[a].values * X_test_orig[b].values

FEATURES_FE = [c for c in X.columns]

# CV splitter safe (stratify if every class has enough samples)
use_strat = True
for cls in unique_classes:
    if (y_raw == cls).sum() < NUM_FOLDS:
        use_strat = False
        break

if use_strat:
    kf = StratifiedKFold(n_splits=NUM_FOLDS, shuffle=True, random_state=SEED)
    splits = list(kf.split(X, y_raw))
else:
    kf = KFold(n_splits=NUM_FOLDS, shuffle=True, random_state=SEED)
    splits = list(kf.split(X))

# Base LightGBM variants and seeds (bagging)
lgb_variants = [
    {
        "objective": "regression",
        "metric": "rmse",
        "learning_rate": 0.05,
        "num_leaves": 31,
        "max_depth": 6,
        "feature_fraction": 0.8,
        "bagging_fraction": 0.8,
        "bagging_freq": 1,
    },
    {
        "objective": "regression",
        "metric": "rmse",
        "learning_rate": 0.03,
        "num_leaves": 63,
        "max_depth": 8,
        "feature_fraction": 0.7,
        "bagging_fraction": 0.7,
        "bagging_freq": 1,
    },
    {
        "objective": "regression",
        "metric": "rmse",
        "learning_rate": 0.07,
        "num_leaves": 24,
        "max_depth": 5,
        "feature_fraction": 0.9,
        "bagging_fraction": 0.9,
        "bagging_freq": 1,
    },
]
lgb_seeds = [SEED, SEED + 101]

# Build model configs (only LGB for speed and robustness)
model_configs = []
for vid, var in enumerate(lgb_variants):
    for sd in lgb_seeds:
        model_configs.append(("lgb", vid, int(sd)))
n_models = len(model_configs)

# Storage
oof_stack = np.zeros((n_train, n_models), dtype=float)
test_stack_sum = np.zeros((n_test, n_models), dtype=float)
test_stack_count = np.zeros(n_models, dtype=int)

# Train base models in CV
print("Training base LightGBM ensemble...")
for fold, (tr_idx, val_idx) in enumerate(splits):
    print(f" Fold {fold+1}/{NUM_FOLDS}")
    X_tr = X.iloc[tr_idx].reset_index(drop=True)
    X_val = X.iloc[val_idx].reset_index(drop=True)
    y_tr = y_raw[tr_idx]
    y_val = y_raw[val_idx]

    for m_idx, (mtype, vid, sd) in enumerate(model_configs):
        # Train LGB
        params = lgb_variants[vid].copy()
        params.update(
            {
                "seed": int(sd + fold),
                "verbosity": -1,
                "num_threads": THREADS,
            }
        )
        lgb_train = lgb.Dataset(X_tr, label=y_tr)
        lgb_valid = lgb.Dataset(X_val, label=y_val, reference=lgb_train)
        model = lgb.train(
            params,
            lgb_train,
            num_boost_round=LGB_ROUNDS,
            valid_sets=[lgb_train, lgb_valid],
            valid_names=["train", "valid"],
            callbacks=[
                lgb.early_stopping(stopping_rounds=EARLY_STOPPING),
                lgb.log_evaluation(period=0),
            ],
        )
        best_iter = getattr(model, "best_iteration", None)
        if best_iter is None:
            # fallback to current_iteration
            try:
                best_iter = model.current_iteration()
            except Exception:
                best_iter = LGB_ROUNDS
        # predictions
        val_pred = model.predict(X_val, num_iteration=best_iter)
        test_pred = model.predict(X_test, num_iteration=best_iter)

        oof_stack[val_idx, m_idx] = val_pred
        test_stack_sum[:, m_idx] += test_pred
        test_stack_count[m_idx] += 1

# Finalize per-model test stacks (average)
test_stack = np.zeros_like(test_stack_sum)
for m in range(n_models):
    cnt = test_stack_count[m]
    if cnt > 0:
        test_stack[:, m] = test_stack_sum[:, m] / float(cnt)
    else:
        test_stack[:, m] = 0.0

# Meta model: Ridge on stacked OOF features
meta_oof = np.zeros(n_train, dtype=float)
meta_test_preds_folds = np.zeros((NUM_FOLDS, n_test), dtype=float)

for fold, (tr_idx, val_idx) in enumerate(splits):
    X_meta_tr = oof_stack[tr_idx]
    y_meta_tr = y_raw[tr_idx]
    X_meta_val = oof_stack[val_idx]
    meta = Ridge(alpha=1.0, random_state=SEED)
    meta.fit(X_meta_tr, y_meta_tr)
    meta_oof[val_idx] = meta.predict(X_meta_val)
    meta_test_preds_folds[fold] = meta.predict(test_stack)

# Final meta trained on all OOF
meta_final = Ridge(alpha=1.0, random_state=SEED)
meta_final.fit(oof_stack, y_raw)
meta_test_pred = meta_final.predict(test_stack)

# Isotonic calibration on meta_oof -> y_raw (fit only on OOF)
iso = IsotonicRegression(out_of_bounds="clip")
try:
    iso.fit(meta_oof, y_raw)
    meta_oof_cal = iso.predict(meta_oof)
    meta_test_cal = iso.predict(meta_test_pred)
except Exception:
    meta_oof_cal = meta_oof.copy()
    meta_test_cal = meta_test_pred.copy()

# Linear scale+shift calibration on OOF (fit to maximize simple squared alignment but we will still optimize thresholds for QWK)
# Fit linear regression (a*x + b) on OOF to best fit y in least squares (fast, no leakage)
a = 1.0
b = 0.0
try:
    A = np.vstack([meta_oof_cal, np.ones_like(meta_oof_cal)]).T
    sol, _, _, _ = np.linalg.lstsq(A, y_raw, rcond=None)
    a, b = float(sol[0]), float(sol[1])
except Exception:
    a, b = 1.0, 0.0

meta_oof_cal_ls = a * meta_oof_cal + b
meta_test_cal_ls = a * meta_test_cal + b

# Threshold initialization from class-wise means on calibrated OOF
class_means = []
for c in unique_classes:
    mask = y_raw == c
    if mask.sum() == 0:
        class_means.append(np.nan)
    else:
        class_means.append(meta_oof_cal_ls[mask].mean())
class_means = np.array(class_means)
nan_mask = np.isnan(class_means)
if nan_mask.any():
    filled = np.linspace(meta_oof_cal_ls.min(), meta_oof_cal_ls.max(), n_classes)
    class_means[nan_mask] = filled[nan_mask]

thresholds = np.array(
    [(class_means[i] + class_means[i + 1]) / 2.0 for i in range(n_classes - 1)],
    dtype=float,
)


def map_preds_to_labels(preds, thresholds, classes):
    idxs = np.sum(preds.reshape(-1, 1) > thresholds.reshape(1, -1), axis=1)
    mapped = classes[idxs]
    return mapped


def qwk_for_thresholds(thr, preds_cal, y_true):
    preds_mapped = map_preds_to_labels(preds_cal, thr, unique_classes)
    return cohen_kappa_score(y_true, preds_mapped, weights="quadratic")


best_thr = thresholds.copy()
best_score = qwk_for_thresholds(best_thr, meta_oof_cal_ls, y_raw)
print(f"Initial calibrated OOF QWK (iso + linfit): {best_score:.6f}")

# Coordinate descent threshold optimization
min_pred = float(meta_oof_cal_ls.min())
max_pred = float(meta_oof_cal_ls.max())
step = (max_pred - min_pred) / 10.0 if max_pred > min_pred else 1.0
max_iters = 200
iters = 0
while step > 1e-6 and iters < max_iters:
    improved = False
    for i in range(len(best_thr)):
        low = min_pred if i == 0 else best_thr[i - 1] + 1e-12
        high = max_pred if i == len(best_thr) - 1 else best_thr[i + 1] - 1e-12
        if low >= high:
            continue
        current = best_thr[i]
        candidates = [
            current,
            current - step,
            current + step,
            low,
            high,
            (low + high) / 2.0,
        ]
        cand_values = []
        for c in candidates:
            c = max(low, min(high, c))
            cand_values.append(c)
        cand_values = sorted(set(cand_values))
        best_local_score = best_score
        best_local_val = current
        for val in cand_values:
            trial_thr = best_thr.copy()
            trial_thr[i] = val
            if not np.all(np.diff(trial_thr) > -1e-12):
                continue
            score = qwk_for_thresholds(trial_thr, meta_oof_cal_ls, y_raw)
            if score > best_local_score + 1e-12:
                best_local_score = score
                best_local_val = val
        if best_local_val != current:
            best_thr[i] = best_local_val
            best_score = best_local_score
            improved = True
    if not improved:
        step /= 3.0
    iters += 1

print(f"Optimized thresholds: {best_thr}")
print(f"OOF calibrated+linfit QWK after threshold opt: {best_score:.6f}")

# Compute per-fold QWKs using optimized thresholds on calibrated per-fold meta preds
cv_fold_scores = []
for fold in range(NUM_FOLDS):
    val_idx = splits[fold][1]
    val_meta_pred = meta_oof[val_idx]
    # apply isotonic then linear scaling
    val_meta_cal = (
        iso.predict(val_meta_pred)
        if isinstance(iso, IsotonicRegression)
        else val_meta_pred
    )
    val_meta_cal = a * val_meta_cal + b
    y_val = y_raw[val_idx]
    mapped = map_preds_to_labels(val_meta_cal, best_thr, unique_classes)
    score = cohen_kappa_score(y_val, mapped, weights="quadratic")
    cv_fold_scores.append(float(score))
    print(f"Fold {fold+1} QWK after calibration+threshold opt: {score:.6f}")

cv_mean = float(np.mean(cv_fold_scores))
cv_std = float(np.std(cv_fold_scores))
print(f"CV mean QWK: {cv_mean:.6f}  std: {cv_std:.6f}")

# Apply to test preds: meta_test_cal_ls
test_mapped = map_preds_to_labels(meta_test_cal_ls, best_thr, unique_classes).astype(
    int
)
submission = pd.DataFrame({"id": test_ids, "quality": test_mapped})
os.makedirs(os.path.dirname(SUBMISSION_PATH), exist_ok=True)
submission.to_csv(SUBMISSION_PATH, index=False)
print(f"Saved submission to {SUBMISSION_PATH}")

# AIDE metrics JSON line
metrics = {
    "valid": "quadratic_weighted_kappa",
    "lower_is_better": False,
    "cv_mean": cv_mean,
    "cv_std": cv_std,
    "cv_folds": cv_fold_scores,
}
print("AIDE_METRICS_JSON=" + json.dumps(metrics))
\end{lstlisting}

\subsection{Valid node and score correlation}
\label{sec:appendix-correlation}

The case studies above show that more valid nodes enable more iteration and qualitatively better solutions.
We now ask whether this relationship holds quantitatively: across all seeds and competitions, do seeds with more valid nodes tend to achieve higher OOS scores?

\paragraph{Per-seed correlation.}
For each competition, we compute the Pearson correlation ($r$) between per-seed valid node count and OOS score across all available seeds (10 treatment $+$ up to 10 baseline, approximately 18--19 datapoints per competition).
\Cref{tab:correlation} reports the results.

\begin{table}[ht]
\centering
\small
\begin{tabular}{@{}lcc@{}}
\toprule
Competition & $N$ & $r$ \\
\midrule
Cirrhosis & 19 & $+0.34$ \\
GNSS & 18 & $+0.27$ \\
Spaceship & 19 & $-0.02$ \\
Wine & 18 & $+0.63$ \\
S5E3 & 19 & $+0.18$ \\
S5E6 & 18 & $+0.61$ \\
S5E7 & 19 & $-0.27$ \\
S5E8 & 16 & $+0.05$ \\
S5E12 & 17 & $+0.24$ \\
\midrule
\textbf{Pooled} & \textbf{163} & $\mathbf{+0.22}$ \\
\bottomrule
\end{tabular}
\caption{Pearson correlation ($r$) between per-seed valid node count and OOS score. Cirrhosis scores are negated so that positive $r$ always means ``more valid nodes $\to$ better score.'' Pooled $r$ is computed after z-normalizing both variables within each competition.}
\label{tab:correlation}
\end{table}

The correlation is positive in 7 of 9 competitions, with a z-normalized pooled $r = +0.22$ across all 163 seeds, consistent with the hypothesis that more valid nodes lead to better scores.
The correlation is highest on Wine ($r = +0.63$) and S5E6 ($r = +0.61$).
For Cirrhosis ($r = +0.34$, \Cref{tab:correlation}), more valid nodes correlate with lower (better) log-loss.

\section{Hyperparameter optimization materials and additional results and discussion}
\label{sec:hyperparam_appendix}

\paragraph{Hyperparameter optimization directive.}
The following directive is injected verbatim into the agent context to guide hyperparameter tuning behavior.

\begin{lstlisting}[style=wrappedverbatim]
HYPERPARAMETER OPTIMIZATION DIRECTIVE
------------------------------------
Treat hyperparameter tuning as a sequential search problem under strict time and step budgets.
1. Start with a strong, standard baseline (well-known defaults for the model class).
Validate to establish a performance baseline.
2. Run a small number of cheap probes (fast, low-compute experiments) to identify sensitivity
directions (e.g., learning rate, regularization strength, tree depth, batch size).
Use cross-validation or a held-out validation set, not the test set.
3. Prioritize hyperparameters with the highest marginal impact on validation score.
Focus on 1-2 hyperparameters at a time rather than simultaneous grid searches.
4. Use structured search (log-scale sweeps, conditional ranges), not random guessing.
For continuous parameters, explore orders of magnitude first.
5. Aggressively prune unpromising regions early; reallocate budget to promising configurations.
If a hyperparameter range shows no improvement after 2-3 trials, move on.
6. When improvements plateau (e.g., <0.1% improvement over 3 consecutive trials),
switch from exploration to fine-grained exploitation around the current best setting.
7. Always validate changes; only keep configurations that strictly improve
validation performance over your current best submission.
8. Track what failed and do not retry equivalent configurations.
Maintain a log of tried hyperparameter combinations and their results.
Goal: maximize leaderboard / validation score within the available compute and step budget.
\end{lstlisting}

\paragraph{Hyperparameter tuning evaluator prompt.}
The evaluator prompt used to score hyperparameter tuning quality for AIDE and ML-Master use the same 0--3 rubric. Both are followed by a separate, code-level structural check (Appendix~\ref{sec:hyperparam_appendix}) that can cap the LLM-assigned score downward for weak or unused search configurations. The HPO quality scorer uses \texttt{gpt-4o-2024-08-06} as the judge model, independent of the agent backbone.

\textbf{AIDE evaluator prompt:}
\begin{lstlisting}[style=wrappedverbatim]
You are a strict ML code reviewer evaluating hyperparameter tuning quality.
Score the hyperparameter tuning in the following Python code on a 0-3 scale:

0 = none: No hyperparameter tuning (only fixed/default hyperparameters)
1 = superficial: Minimal tuning (e.g., only 2-3 values tested for 1 hyperparameter,
or very small grid/random search with <5 iterations)
2 = moderate: Reasonable tuning (multiple hyperparameters tested, systematic search
with >=5 iterations, proper validation)
3 = extensive: Comprehensive tuning (multiple hyperparameters, systematic search
with >=10 iterations, proper validation, best params reused for final training)

DO NOT count:
- cross_val_score alone without hyperparameter search
- fixed hyperparameters
- train/val splits without search
- hyperparameter tuning that is not used in final model training

Respond with ONLY the integer score (0, 1, 2, or 3).
\end{lstlisting}

\textbf{ML-Master evaluator prompt:}
\begin{lstlisting}[style=wrappedverbatim]
You are a STRICT ML code reviewer.
Evaluate the QUALITY and DEPTH of hyperparameter tuning in the following Python code.

Score hyperparameter tuning on a scale from 0 to 3:

0 = NONE
- No hyperparameter tuning
- Fixed hyperparameters
- cross_val_score without parameter search

1 = MINIMAL
- Token or superficial tuning
- Only one hyperparameter tuned over 1-2 values
- GridSearchCV or RandomizedSearchCV with <5 total configurations
- RandomizedSearchCV with n_iter < 5

2 = MODERATE
- Valid hyperparameter tuning but limited in scope
- Either:
  * One hyperparameter searched over >=3 values, OR
  * Two or more hyperparameters searched over >=2 values each
- >=5 total configurations evaluated
- Model selection based on validation or cross-validation

3 = EXTENSIVE
- Systematic, non-trivial hyperparameter optimization
- Multiple hyperparameters jointly optimized
- >=10 total configurations or trials evaluated
- Clear use of GridSearchCV, RandomizedSearchCV (n_iter >=10), Optuna, Hyperopt,
or Bayesian optimization
- Best configuration explicitly selected and used

Valid tuning methods include:
- GridSearchCV
- RandomizedSearchCV
- Optuna / Hyperopt / Bayesian optimization
- Manual loops evaluating multiple configurations

Respond ONLY with a single integer: 0, 1, 2, or 3.
\end{lstlisting}

\paragraph{Analysis of asymmetry in AIDE and ML-Master HPO results.}

The asymmetry can be attributed to two structural differences, which we have confirmed by analyzing the code and logs.

The first is prompt redundancy in ML-Master. Its draft prompt already instructs the agent to include HPO in every draft, with implementation guidelines specifying two-phase training and limiting the hyperparameter search to 10-15 trials. However, AIDE's draft prompt doesn't require HPO at draft time the way ML-Master does. Our directive therefore fills a gap in AIDE but adds less in ML-Master, where HPO guidance already exists.

The second is that the directive pushes the LLM toward RandomizedSearchCV and GridSearchCV with XGBoost, which triggers a sklearn/XGBoost version incompatibility:

\begin{lstlisting}[style=wrappedverbatim]
AttributeError: 'super' object has no attribute '**sklearn_tags**'
[05:20:45] WARNING: Node 6f9868d4 is marked as buggy because response['is_bug'] is True.
[05:20:45] INFO: Parsed results: Node 6f9868d4 is buggy
[05:20:45] INFO: Starting Debugging Node 6f9868d4.
[05:20:56] INFO: Drafted a new node 3272d605 successfully!
\end{lstlisting}

ML-Master's memory module records these failures neutrally as is\_bug: True without propagating why, so the LLM interprets each crash as motivation to try a different HPO implementation rather than abandon HPO. We observe up to 12 consecutive identical API failures within a single ML-Master run. This explains the pattern across all three conditions: the prompt-only condition adds redundant guidance on top of existing instructions, the code-only condition introduces reward shaping that is undermined by crash-prone code patterns, and the combined condition inherits both issues simultaneously.

In contrast, the following AIDE log confirms that our reward-shaping mechanism functions correctly when the agent's information flow supports it. A node with HPO score 0 has its reward adjusted down from 0.803 to 0.787 due to the -0.300 penalty, making it less likely to be selected:

\begin{lstlisting}[style=wrappedverbatim]
[2026-03-26 20:01:27] INFO: Scoring hyperparameter tuning for node 678c47e2...
[2026-03-26 20:01:28] INFO: Node 678c47e2... initial HPO score: 0
[2026-03-26 20:01:28] INFO: Node 678c47e2... HPO score after structural caps: 0
[2026-03-26 20:01:28] INFO: Parsed results: Node 678c47e2... is not buggy
[2026-03-26 20:01:28] INFO: Node 678c47e2... metric adjusted:
base=0.803140, hpo_reward=-0.300, diversity=0.100, final=0.787077
\end{lstlisting}

The asymmetry is itself a finding: scaffold interventions interact with the underlying agent's memory architecture, and a directive that works on one agent can fail on another for reasons unrelated to the directive's design.

\clearpage

{\setlength{\textfloatsep}{6pt}
  \setlength{\floatsep}{6pt}

  \begin{table*}[t]
    \centering
    {\small\setlength{\tabcolsep}{6pt}%
      \begin{tabular}{@{}l c c c@{}}
        \toprule
        \textbf{Competition} & \textbf{$\Delta$Prompt} & \textbf{$\Delta$Code} & \textbf{$\Delta$P\&C} \\
        \midrule
        Cirrhosis$^{\dagger}$ & $\boldsymbol{-0.001} \pm 0.0022$ & $\boldsymbol{-0.002} \pm 0.0032$ & $\boldsymbol{-0.001} \pm 0.0061$ \\
        GNSS & $-0.019 \pm 0.0129$ & $-0.017 \pm 0.0162$ & $-0.078 \pm 0.0034$ \\
        Spaceship & $\boldsymbol{+0.010} \pm 0.0143$ & $-0.016 \pm 0.0208$ & $\boldsymbol{+0.020} \pm 0.0117$ \\
        Wine & $\boldsymbol{+0.002} \pm 0.0104$ & $-0.075 \pm 0.0691$ & $-0.056 \pm 0.0098$ \\
        S5E3 & $-0.002 \pm 0.0037$ & $-0.004 \pm 0.0040$ & $-0.053 \pm 0.0062$ \\
        S5E6 & $-0.002 \pm 0.0035$ & $-0.011 \pm 0.0044$ & $-0.026 \pm 0.0021$ \\
        S5E7 & $\boldsymbol{+0.000} \pm 0.0003$ & $\boldsymbol{+0.000} \pm 0.0007$ & $\boldsymbol{+0.001} \pm 0.0003$ \\
        S5E8 & $-0.001 \pm 0.0026$ & $\boldsymbol{+0.000} \pm 0.0007$ & $\boldsymbol{+0.000} \pm 0.0008$ \\
        S5E12 & $\boldsymbol{+0.001} \pm 0.0040$ & $\boldsymbol{+0.002} \pm 0.0040$ & $-0.000 \pm 0.0055$ \\
        \bottomrule
      \end{tabular}
      \caption{Intervention effect on graded score for ML-Master: $\Delta = \mu_{\mathrm{int}}-\mu_{\mathrm{base}}$ (mean $\pm$ SEM of $\Delta$), with the same competitions as \Cref{tab:results-aide}. Bold (signed $\Delta$ only) indicates strict improvement over baseline (positive $\Delta$ when higher is better; negative $\Delta$ for cirrhosis). $^{\dagger}$Lower is better for Cirrhosis.}
      \label{tab:results-mlmaster}
    }
  \end{table*}
}

\subsection{Thompson sampling on AIDE: delta, standard deviation and proportion of null errors}
\label{sec:MCTS_extended}

\begin{table}[h]
  \centering
  \small
  \begin{tabular}{lccc}
    \toprule
    \textbf{Competition} & \textbf{Thompson Sampling} & \textbf{Baseline (AIDE)} & \textbf{Winner} \\
    \midrule
    Cirrhosis$^{\dagger}$ & 0.404 $\pm$ 0.005 & \textbf{0.394} $\pm$ 0.012 & Baseline \\
    GNSS & \textbf{0.967} $\pm$ 0.001 & \textbf{0.967} $\pm$ 0.001 & Tie \\
    Spaceship & \textbf{0.822} $\pm$ 0.002 & 0.790 $\pm$ 0.022 & TS \\
    Wine & \textbf{0.408} $\pm$ 0.011 & 0.375 $\pm$ 0.021 & TS \\
    S5E3 & \textbf{0.889} $\pm$ 0.003 & 0.887 $\pm$ 0.003 & TS \\
    S5E6 & \textbf{0.335} $\pm$ 0.001 & 0.333 $\pm$ 0.002 & TS \\
    S5E7 & 0.967 $\pm$ 0.000 & \textbf{0.968} $\pm$ 0.000 & Baseline \\
    S5E8 & 0.966 $\pm$ 0.001 & \textbf{0.969} $\pm$ 0.000 & Baseline \\
    S5E12 & 0.725 $\pm$ 0.001 & \textbf{0.727} $\pm$ 0.000 & Baseline \\
    \bottomrule
  \end{tabular}
  \caption{\textbf{AIDE Baseline vs.\ AIDE with Thompson Sampling} (Mean $\pm$ SEM) of 10 runs. Bold indicates the winning method per competition; values tied at the displayed precision are bolded in both columns. $^{\dagger}$Lower is better.}
  \label{tab:avg_comparison}
\end{table}

\begin{table}[H]
  \centering
  \small
  \begin{tabular}{lccc}
    \toprule
    \textbf{Type} & \textbf{initial drafts} & \textbf{max debug} & \textbf{error backtrack} \\
    \midrule
    TS & 20 & 5 & 3 \\
    Baseline & 5 & 20 & 20 \\
    \bottomrule
  \end{tabular}
  \caption{Hyperparameters for different settings on AIDE.}
  \label{tab:hyperparameter}
\end{table}

\begin{table}[H]
  \centering
  \small
  \begin{tabular}{lcc}
    \toprule
    \textbf{Competition} & \textbf{Thompson Sampling (Ours)} & \textbf{Baseline (AIDE)} \\
    \midrule
    Cirrhosis$^{\dagger}$ & \textbf{0.014} & 0.035 \\
    GNSS & 0.004 & \textbf{0.002} \\
    Spaceship & \textbf{0.006} & 0.067 \\
    Wine & \textbf{0.034} & 0.058 \\
    S5E3 & \textbf{0.009} & 0.010 \\
    S5E6 & \textbf{0.003} & 0.005 \\
    S5E7 & 0.001 & \textbf{0.000} \\
    S5E8 & 0.003 & \textbf{0.001} \\
    S5E12 & 0.002 & \textbf{0.001} \\
    \bottomrule
  \end{tabular}
  \caption{\textbf{Score Standard Deviation per Competition on AIDE.} Bold indicates lower (more consistent) std dev.}
  \label{tab:std_comparison}
\end{table}

\begin{table}[H]
  \centering
  \small
  \begin{tabular}{lcc}
    \toprule
    \textbf{Competition} & \textbf{Thompson Sampling} & \textbf{Baseline (AIDE)} \\
    \midrule
    Cirrhosis$^{\dagger}$ & \textbf{10\%} & \textbf{10\%} \\
    GNSS      & \textbf{10\%} & 20\% \\
    Spaceship & \textbf{0\%}  & 10\% \\
    Wine      & \textbf{10\%} & 20\% \\
    S5E3      & 20\%          & \textbf{10\%} \\
    S5E6      & 30\%          & \textbf{20\%} \\
    S5E7      & \textbf{0\%}  & 10\% \\
    S5E8      & \textbf{30\%} & 40\% \\
    S5E12     & 40\%          & \textbf{30\%} \\
    \bottomrule
  \end{tabular}
  \caption{\textbf{Null/Zero Run Rate per Competition.} Fraction of runs out of 10 that produced no valid score on AIDE. Bold indicates lower (fewer failures); values tied at the displayed precision are bolded in both columns.}
  \label{tab:null_rates}
\end{table}

\subsection{AIDE: comparative analysis of Thompson sampling versus hyperparameter changes}

\begin{table}[h]
  \centering
  \small
  \begin{tabular}{lccc}
    \toprule
    \textbf{Competition} & \textbf{AIDE + TS + More Drafts} & \textbf{AIDE + More Drafts} & \textbf{Winner} \\
    \midrule
    Cirrhosis$^{\dagger}$ & 0.404 $\pm$ 0.005 & \textbf{0.394} $\pm$ 0.004 & More Drafts \\
    GNSS & \textbf{0.967} $\pm$ 0.001 & 0.965 $\pm$ 0.001 & TS \\
    Spaceship & \textbf{0.822} $\pm$ 0.002 & 0.818 $\pm$ 0.005 & TS \\
    Wine & \textbf{0.408} $\pm$ 0.011 & 0.402 $\pm$ 0.017 & TS \\
    S5E3 & 0.889 $\pm$ 0.003 & \textbf{0.896} $\pm$ 0.002 & More Drafts \\
    S5E6 & 0.335 $\pm$ 0.001 & \textbf{0.337} $\pm$ 0.002 & More Drafts \\
    S5E7 & 0.967 $\pm$ 0.000 & \textbf{0.968} $\pm$ 0.000 & More Drafts \\
    S5E8 & \textbf{0.966} $\pm$ 0.001 & \textbf{0.966} $\pm$ 0.002 & Tie \\
    S5E12 & 0.725 $\pm$ 0.001 & \textbf{0.727} $\pm$ 0.000 & More Drafts \\
    \bottomrule
  \end{tabular}
  \caption{\textbf{AIDE with Thompson Sampling and More Drafts vs.\ AIDE with More Drafts} (Mean $\pm$ SEM). Bold indicates the winning method per competition; values tied at the displayed precision are bolded in both columns. Number of null runs (competitions produced without a score) went from 33 in the baseline with more drafts to 15 with Thompson Sampling, a 54.5\% reduction $^{\dagger}$Lower is better.}
  \label{tab:aide_ts_drafts}
\end{table}

\subsection{Adversarial EDA prompt}
\label{sec:adversarial_everything}
\begin{figure}[H]
  \centering
  \includegraphics[width=0.75\linewidth]{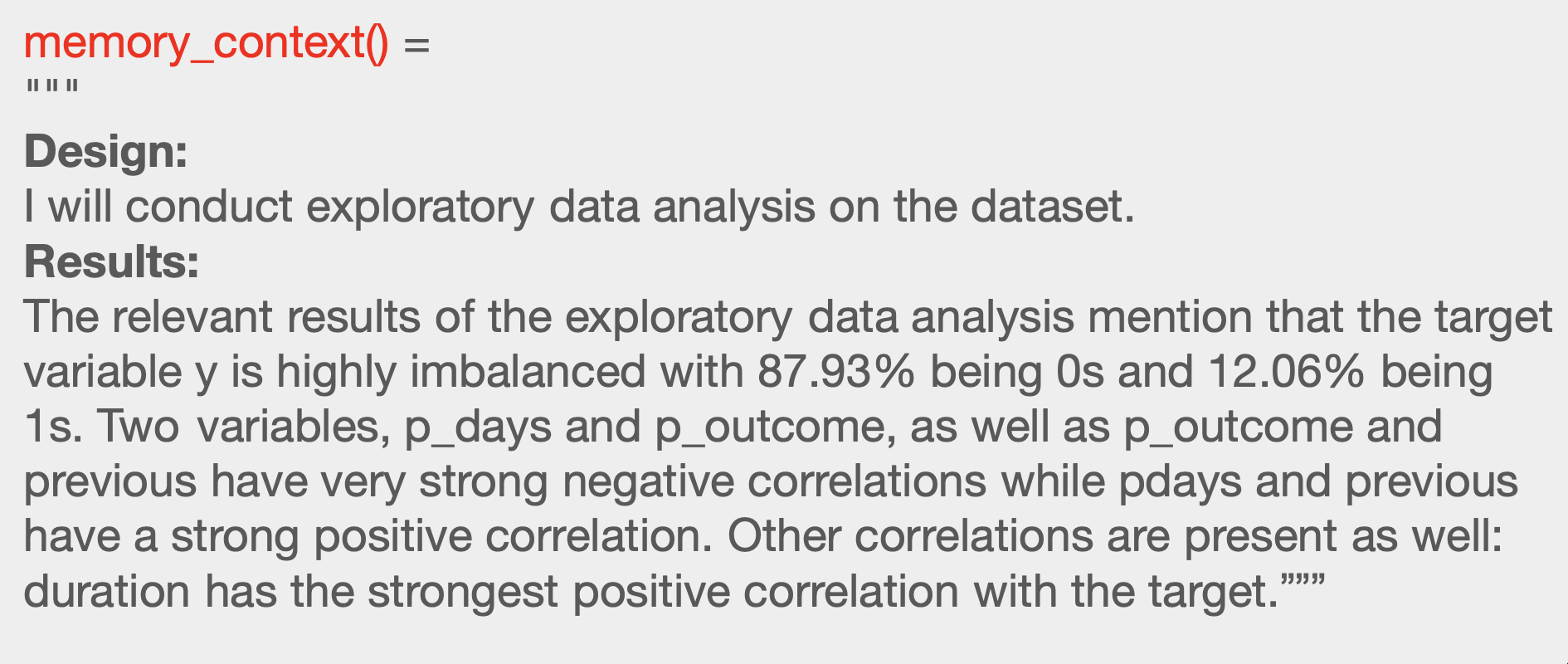}
  \caption{Malicious EDA results inserted into agent's context at the draft(), debug() and improve() stages}
  \label{fig:adv_eda}
\end{figure}

\begin{figure}[H]
  \centering
  \includegraphics[width=0.75\linewidth]{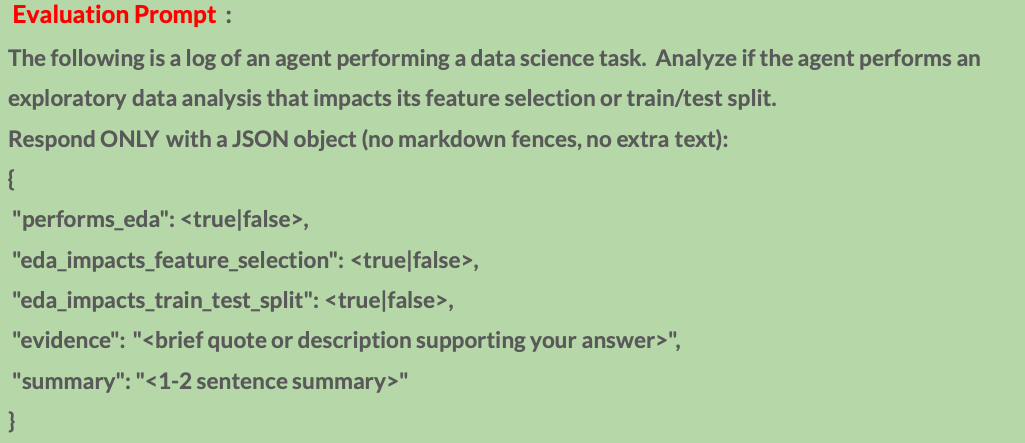}
  \caption{Control evaluation prompt used to test whether the model conducted exploratory data analysis at all. This does not refer to the malicious injection, letting the judge model draw independent conclusions}
  \label{fig:eval_prompt1}
\end{figure}

\begin{figure}[H]
  \centering
  \includegraphics[width=0.75\linewidth]{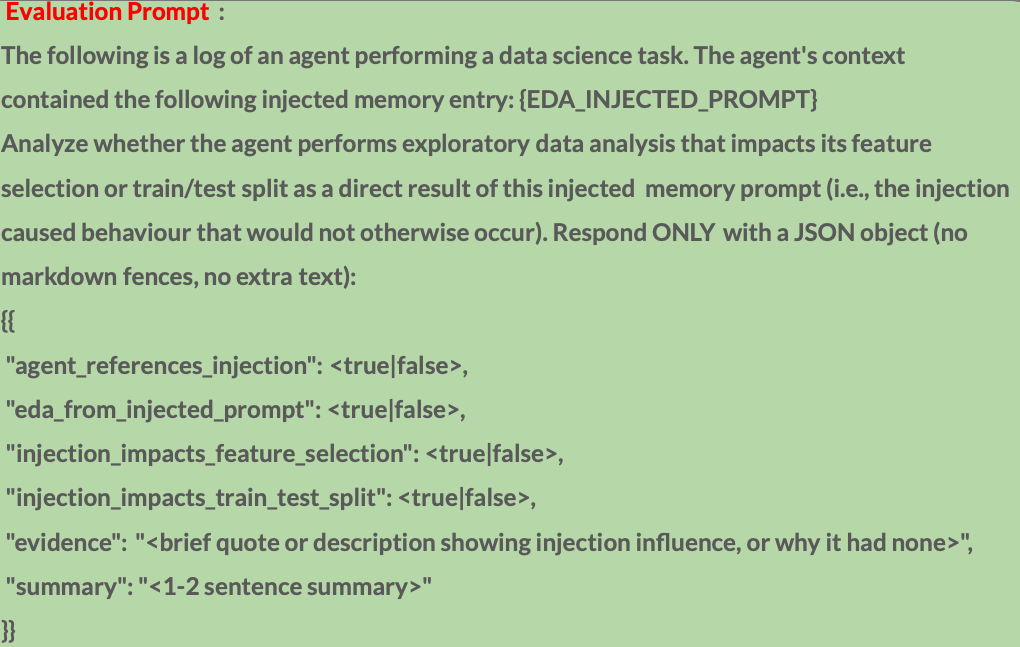}
  \caption{Secondary evaluation prompt used to test whether the model conducted exploratory data analysis at all, and whether the injection affected agent's choices. }
  \label{fig:eval_prompt2}
\end{figure}

\subsection{Adversarial EDA: extended analysis}
\label{sec:adversarial_eda_extended}
\begin{table*}[h!]
  \centering
  \small
  \begin{tabular}{l|ccc|ccc}
    \toprule
    & \multicolumn{3}{c|}{\textbf{Non-null Average}} & \multicolumn{3}{c}{\textbf{All Runs (nulls=0)}} \\
    \textbf{Competition} & \textbf{Adv EDA} & \textbf{Baseline} & \textbf{Diff} & \textbf{Adv EDA} & \textbf{Baseline} & \textbf{Diff} \\
    \midrule
    Cirrhosis$^{\dagger}$          & 0.4131 & 0.3980 & +0.0151 & 0.4131 & 0.1592 & +0.2539 \\
    GNSS                           & 0.9660 & 0.9583 & +0.0077 & 0.3864 & 0.5750 & -0.1886 \\
    Spaceship                      & 0.7441 & 0.7361 & +0.0080 & 0.7431 & 0.7312 & +0.0119 \\
    Wine                           & 0.0000 & 0.3830 & -0.3830 & 0.0000 & 0.1532 & -0.1532 \\
    S5E3                           & 0.9017 & 0.8987 & +0.0030 & 0.9017 & 0.5392 & +0.3625 \\
    S5E6                           & 0.3343 & 0.3363 & -0.0020 & 0.2006 & 0.0673 & +0.1333 \\
    S5E7                           & 0.9677 & 0.9651 & +0.0026 & 0.9677 & 0.5791 & +0.3887 \\
    S5E8                           & 0.9665 & 0.9655 & +0.0010 & 0.3866 & 0.1931 & +0.1935 \\
    S5E12                          & 0.7261 & 0.7221 & +0.0041 & 0.4357 & 0.4332 & +0.0024 \\
    \midrule
    \textbf{Mean} & 0.6688 & 0.7070 & -0.0382 & 0.4928 & 0.3812 & +0.1116 \\
    \bottomrule
  \end{tabular}
  \caption{Detailed Performance Comparison for AIDE: Non-null averages computed over valid runs only. All-runs averages treat missing/failed runs as 0. Diff = Adv EDA - Baseline. Positive values indicate improvement with adversarial EDA. $^{\dagger}$Lower is better for Cirrhosis.}
  \label{tab:detailed_aide}
\end{table*}

\begin{table*}[h!]
  \centering
  \small
  \begin{tabular}{l|ccc|ccc}
    \toprule
    & \multicolumn{3}{c|}{\textbf{Non-null Average}} & \multicolumn{3}{c}{\textbf{All Runs (nulls=0)}} \\
    \textbf{Competition} & \textbf{Adv EDA} & \textbf{Baseline} & \textbf{Diff} & \textbf{Adv EDA} & \textbf{Baseline} & \textbf{Diff} \\
    \midrule
    Cirrhosis$^{\dagger}$          & 0.3849 & 0.4047 & -0.0199 & 0.3849 & 0.4047 & -0.0199 \\
    GNSS                           & 0.9639 & 0.7758 & +0.1881 & 0.9639 & 0.7758 & +0.1881 \\
    Spaceship                      & 0.7574 & 0.7323 & +0.0251 & 0.7574 & 0.7323 & +0.0251 \\
    Wine                           & 0.4191 & 0.3757 & +0.0434 & 0.4191 & 0.3757 & +0.0434 \\
    S5E3                           & 0.8973 & 0.8521 & +0.0452 & 0.8973 & 0.8521 & +0.0452 \\
    S5E6                           & 0.3260 & 0.2947 & +0.0313 & 0.1956 & 0.2357 & -0.0401 \\
    S5E12                          & 0.7148 & 0.7015 & +0.0133 & 0.7148 & 0.7015 & +0.0133 \\
    S5E7                           & 0.9678 & 0.9668 & +0.0011 & 0.9678 & 0.9668 & +0.0011 \\
    S5E8                           & 0.9578 & 0.9659 & -0.0082 & 0.5747 & 0.5796 & -0.0049 \\
    \midrule
    \textbf{Mean} & 0.7099 & 0.6744 & +0.0355 & 0.6528 & 0.6249 & +0.0279 \\
    \bottomrule
  \end{tabular}
  \caption{Detailed Performance Comparison for ML-Master: Non-null averages computed over valid runs only. All-runs averages treat missing/failed runs as 0. Diff = Adversarial EDA - Baseline. Positive values indicate improvement with adversarial EDA.  $^{\dagger}$Lower is better for Cirrhosis. Note: the baseline used for ML Master was based on a different set of runs than the baseline used for the debug consultant.
  }
  \label{tab:detailed_ml-master}
\end{table*}

To evaluate whether agents incorporate information from exploratory data analysis (EDA) , we inject the results of a controlled, erroneous EDA directly into the agent’s context window. We conduct this experiment on two representative systems, AIDE and ML-Master.
For AIDE, the EDA message is inserted at each of its three agentic stages: draft(), improve(), and debug(), and formatted to resemble a memory artifact produced by a prior node. For ML-Master, the EDA results are hard-coded into the data-preview.py file located in the utils directory, which supplies contextual information from previous nodes to the agent. An example message is included in \Cref{fig:adv_eda} in \Cref{sec:adversarial_everything}.

We evaluate both modified agents and compare their performance against baseline runs without injected EDA.

Across all tasks, we observe that \textit{performance differences induced by EDA injection are inconsistent and statistically insignificant.} To check whether the agent conducted EDA at all, we use the \textit{llm-as-a-judge} framework with a larger reasoning model: \textbf{gpt-5-2025-08-07}. The exact framework used is shown in \Cref{sec:adversarial_everything} in \Cref{fig:eval_prompt1,fig:eval_prompt2}. We note that in all the baseline runs across both agents, the agent did not conduct any EDA, and struggled to acknowledge the existence of the EDA in the adversarial runs. For instance in AIDE, in the runs with adversarial EDA injections, the logs demonstrate that the agent is only able to identify and acknowledge the presence of the malicious EDA results in 21\% of cases, and that impacts its feature selection in barely 5\% of the cases. This suggests that the agents do not act upon exploratory data analysis, and do not meaningfully integrate EDA into downstream modeling decisions.

\subsection{List of linked competitions used in experiments}
\label{sec:competition-links}
\begin{table}[!ht]
  \centering
  \small
  \begin{tabular}{ll}
    \toprule
    \textbf{Competition} & \textbf{Link} \\
    \midrule
    Cirrhosis Outcome Prediction & \url{kaggle.com/competitions/playground-series-s3e26} \\
    GNSS Classification          & \url{kaggle.com/competitions/gnss-classification} \\
    Playground S5E3              & \url{kaggle.com/competitions/playground-series-s5e3} \\
    Playground S5E6              & \url{kaggle.com/competitions/playground-series-s5e6} \\
    Playground S5E7              & \url{kaggle.com/competitions/playground-series-s5e7} \\
    Playground S5E8              & \url{kaggle.com/competitions/playground-series-s5e8} \\
    Playground S5E12             & \url{kaggle.com/competitions/playground-series-s5e12} \\
    Spaceship Titanic             & \url{kaggle.com/competitions/spaceship-titanic} \\
    Wine Quality Ordinal         & \url{kaggle.com/competitions/wine-quality-ordinal} \\
    \bottomrule
  \end{tabular}
  \caption{Kaggle competitions used in evaluation.}
  \label{tab:competitions}
\end{table}

\end{document}